\documentclass{article}
\usepackage[T1]{fontenc}
\PassOptionsToPackage{compatV3}{fancyhdr}
\usepackage[preprint]{colm2026_conference}
\usepackage{microtype}
\usepackage{url}
\usepackage{xcolor}
\usepackage{booktabs}
\usepackage{array}
\usepackage{graphicx}
\usepackage{placeins}
\usepackage{amsmath,amssymb,amsthm,bm}
\usepackage{longtable}
\usepackage{hyperref}
\definecolor{darkblue}{rgb}{0,0,0.5}
\hypersetup{colorlinks=true,citecolor=darkblue,linkcolor=darkblue,urlcolor=darkblue}
\newcommand{\majk}[1]{\mathrm{maj}@#1}
\newcommand{\Exp}{\mathbb{E}}

\newcommand{\numQwenFFNFiveMaxResidual}{0.94}
\newcommand{\numGemmaTwentyMaxResidual}{3.25}
\newcommand{\numRankGemmaTwentyStable}{2}

\newcommand{\numChunkGemmaTwenty}{2}

\newcommand{\tabAllocationExamples}{%
\begingroup\setlength{\tabcolsep}{3pt}
\begin{tabular*}{\textwidth}{@{\extracolsep{\fill}}lrrrrl@{}}
\toprule
\textbf{Model} & \textbf{Uniform} & \textbf{Original} & \textbf{\shortstack{Mean-adjusted\\original}} & \textbf{Pruned} & \textbf{$\Delta_M$ [95\% range]} \\
\midrule
Gemma-4-26B-A4B-it E20 & 77.72 & 78.62 & 80.34 & 82.57 & +2.23 [+0.01, +4.20] \\
Qwen3.6-35B-A3B E75 & 81.14 & 67.14 & 77.79 & 85.70 & +7.91 [+2.63, +11.85] \\
GLM-4.7-Flash E50 & 69.29 & 64.30 & 70.04 & 74.48 & +4.44 [+0.02, +10.02] \\
Qwen3.6-35B-A3B FFN5 & 89.85 & 93.56 & 93.76 & 93.70 & -0.07 [-0.32, +0.53] \\
\bottomrule
\end{tabular*}\endgroup}

\newcommand{\numAllocationB}{1000}

\newcommand{\mismatchQwenFFNFiveIIDSup}{5.76}
\newcommand{\mismatchGemmaTwentyIIDSup}{25.50}
\newcommand{\mismatchGemmaTwentyChunkSup}{6.36}

\newcommand{\allocationControlExamples}{%
\begingroup\setlength{\tabcolsep}{3pt}
\begin{tabular*}{\textwidth}{@{\extracolsep{\fill}}lrrrrr@{}}\toprule
\textbf{Model} & \textbf{Cal} & \textbf{Next $\Delta_M$} & \textbf{Next $\Delta_R$} & \textbf{Incl. $\Delta_M$} & \textbf{Incl. $\Delta_R$} \\
\midrule
Gemma-4-26B-A4B-it E20 & 25/30 & +2.23 & +1.97 & -0.29 & +0.00 \\
Qwen3.6-35B-A3B E75 & 27/30 & +7.91 & +5.29 & -1.00 & +0.00 \\
GLM-4.7-Flash E50 & 24/30 & +4.44 & +2.37 & -2.24 & -0.50 \\
Qwen3.6-35B-A3B FFN5 & 28/30 & -0.07 & -0.11 & -0.00 & +0.00 \\
\bottomrule\end{tabular*}\endgroup}

\newcommand{\allocationStoppingExamples}{%
\begingroup\setlength{\tabcolsep}{3pt}
\begin{tabular*}{\textwidth}{@{\extracolsep{\fill}}lrrrrrr@{}}\toprule
\textbf{Model} & \textbf{Cal} & \textbf{Used/limit} & \textbf{Unrestricted P} & \textbf{P=U} & \textbf{O} & \textbf{M} \\
\midrule
Gemma-4-26B-A4B-it E20 & 25/30 & 1920/1920 & 85.95 & 86.58 & 86.37 & 86.43 \\
Qwen3.6-35B-A3B E75 & 27/30 & 1920/1920 & 91.08 & 93.15 & 93.15 & 93.15 \\
GLM-4.7-Flash E50 & 24/30 & 1920/1920 & 81.49 & 85.75 & 85.50 & 86.25 \\
Qwen3.6-35B-A3B FFN5 & 28/30 & 1920/1920 & 94.26 & 95.12 & 95.12 & 95.12 \\
\bottomrule\end{tabular*}\endgroup}

\title{Rethinking LLM Pruning via Test-Time Scaling}
\author{Mingyang Song, Mao Zheng \\
Foundation Model Department, Tencent, China \\
\texttt{nickmysong@tencent.com}}

\fancypagestyle{papertitle}{%
  \fancyhead[L]{Work in Progress}%
}

\begin{document}
\maketitle
\fancyhead[L]{}
\thispagestyle{papertitle}

\begin{abstract}
Structured pruning reduces the size of large language models, often at
the cost of lower accuracy. Whether this decline reflects diminished
reasoning capacity or greater difficulty in sampling correct solutions
is less clear. We examine this question across expert, FFN and layer
pruning on four mathematical and coding benchmarks, comparing original
and pruned models through pass@$k$ at matched sampling budgets. Even
after substantial pruning, some models match or exceed the original
model's reasoning coverage despite lower pass@1, and solve problems left
unsolved in the original samples. Retaining broad coverage, however,
does not mean that correct solutions remain easy to obtain. Under severe
pruning, diverse generated chain-of-thought (CoT) traces can coexist with
increased repetition and less reliable generation. As correct answers
become harder to sample, the additional attempts can sharply increase
total token use even when individual responses become shorter. This
separation between coverage and sampling efficiency also clarifies expert-selection gains,
as methods with markedly different pass@1 can solve the same problems at
large budgets yet require different numbers of attempts to reach a
common coverage target. Pruning should therefore be assessed by both the
range of problems a model can still solve and the sampling cost of
obtaining correct solutions.

\end{abstract}

\section{Introduction}
\label{sec:intro}

Structured pruning reduces model size by removing entire experts, FFN
channels or layers~\citep{lu2024expertsparsity,song2026razor,ma2023llmpruner,men2024shortgpt}.
Removing these components can lower accuracy, but the size of that
decline does not by itself tell us how much reasoning capacity remains.
Some problems may become unsolvable, whereas others may still admit
correct solutions that the pruned model generates less often. Both
changes reduce single-attempt accuracy, yet they imply different
prospects for recovering performance through additional sampling.

To examine whether correct solutions remain accessible after pruning,
we compare original and pruned models through repeated sampling,
following coverage analyses of reinforcement learning and layer
pruning~\citep{yue2025rlvr,wang2026layerpruning}. We measure
\emph{reasoning coverage} with pass@$k$, the fraction of problems with at
least one successful response in $k$ attempts, and match sampling
budgets across models~\citep{chen2021codex,brown2024monkeys}.
Tracking coverage as $k$ increases reveals how readily each model
produces successes, while comparing the identities of solved problems
shows whether it retains the original model's successes or reaches
different ones. Because reaching a given coverage can require many
attempts, we also measure the attempts and generated tokens needed to
reach a common coverage target. These comparisons probe reasoning
capacity at finite budgets, without treating an unobserved success as
proof that a problem is unsolvable.

We study FFN and layer pruning on three MoE and four dense backbones,
and expert pruning on the MoE models, without recovery training.
Across these comparisons, we examine how retained coverage depends on
the pruning structure, task and source model. Distinguishing total from
active parameters helps interpret differences between pruning structures
at similar model sizes. At the problem level, we ask whether the original
model's response length helps characterize vulnerability beyond what
its accuracy reveals. Where coverage survives
despite lower accuracy, we examine generation behavior to understand
how obtaining correct answers becomes more costly. This distinction
between retained coverage and sampling cost also guides our comparisons of expert-selection criteria and
combinations of pruning structures, since higher accuracy may reflect
easier sampling without a broader observed range of solvable problems.
Our main findings are as follows.

\noindent\textcolor{darkblue}{$\blacklozenge$\enspace\textbf{Pruning can preserve or expand reasoning coverage.}}
At matched sampling budgets, some pruned models retain the original
model's solved problems and reveal previously unobserved successes.
This preservation depends on the pruning structure as well as the task
and backbone. In comparisons at similar total model sizes, expert
pruning retains more coverage than FFN pruning while also preserving
more active parameters, showing why total size alone is insufficient to
characterize the effect of pruning (Section~\ref{sec:comparisons}).

\noindent\textcolor{darkblue}{$\blacklozenge$\enspace\textbf{Pruning can increase the diversity of generated CoT traces, but severe pruning impairs generation reliability.}}
Greater variation across samples can coexist with more repetition within
responses and disrupted termination, so diverse CoT traces do not by
themselves indicate reliable generation. When correct answers become
rare under severe pruning, the additional attempts can outweigh savings
from shorter responses, sharply increasing the total tokens needed to
reach a coverage target (Section~\ref{sec:process}).

\noindent\textcolor{darkblue}{$\blacklozenge$\enspace\textbf{Original accuracy alone does not characterize pruning robustness.}}
Even problems with similar original success frequencies can suffer
different accuracy losses after pruning. Their original response
lengths provide additional information about this vulnerability, although
the strength of the association depends on the task and pruning
structure. Consistently solving a problem before pruning therefore does
not guarantee that it will remain easy to solve afterward
(Section~\ref{sec:task-demands}).

\noindent\textcolor{darkblue}{$\blacklozenge$\enspace\textbf{Expert-selection gains can reflect improved sampling efficiency without broader observed coverage.}}
At a fixed backbone and pruning ratio, methods with substantially
different pass@1 can recover the same observed solved set at larger
budgets. Their difference lies in how readily they produce these
successes, which affects the attempts and tokens needed to reach a
common coverage target. In these comparisons, the advantage is therefore
one of sampling efficiency rather than an expansion of the observed
range of solvable problems (Section~\ref{sec:method-comparison}).

\noindent\textcolor{darkblue}{$\blacklozenge$\enspace\textbf{Combining pruning across structures does not necessarily improve the trade-off between model size and reasoning capacity.}}
A tested combination of expert, FFN and layer pruning uses fewer active
parameters than expert-only pruning at a similar total size, but retains
less coverage. Judging compression by total size alone would overlook
this difference in both active parameters and retained coverage.
Assessing the benefit of combined pruning therefore requires considering
both parameter counts alongside the range of problems that remain solvable
(Section~\ref{sec:mixed}).

\section{Experimental Setup}
\label{sec:setup}

\subsection{Models and Pruning Comparisons}
\label{sec:design}

We compare
Qwen3.6-35B-A3B\footnote{\url{https://huggingface.co/Qwen/Qwen3.6-35B-A3B}},
GLM-4.7-Flash\footnote{\url{https://huggingface.co/zai-org/GLM-4.7-Flash}} and
Gemma-4-26B-A4B-it~\citep{gemmateam2026gemma4} before and after pruning. All Qwen3.6-35B-A3B
results use the post-trained release.
FFN and layer pruning on four dense backbones test how coverage
retention depends on model architecture
(Appendix~\ref{app:coverage}).

We use RAZOR for expert selection~\citep{song2026razor},
Taylor saliency for FFN channels~\citep{ma2023llmpruner}, and a
perplexity-greedy layer selector adapted from iterative block
pruning~\citep{song2024sleb}.
All three pruning methods use RAZOR's razorcal dataset for calibration,
but differ in how they score units and how much calibration data they use. We apply no
recovery training. Base and Instruct comparisons test dependence on the
source model's training stage, while two combinations of pruning axes
test the trade-off between model size and coverage.
Pruning ratios specify the fraction of structural units removed along
each axis; equal ratios need not remove equal numbers of parameters or
FLOPs. Appendix~\ref{app:operators} defines these ratios and compares
the resulting parameter reductions.

\begin{figure}[!t]
\centering
\includegraphics[width=\textwidth]{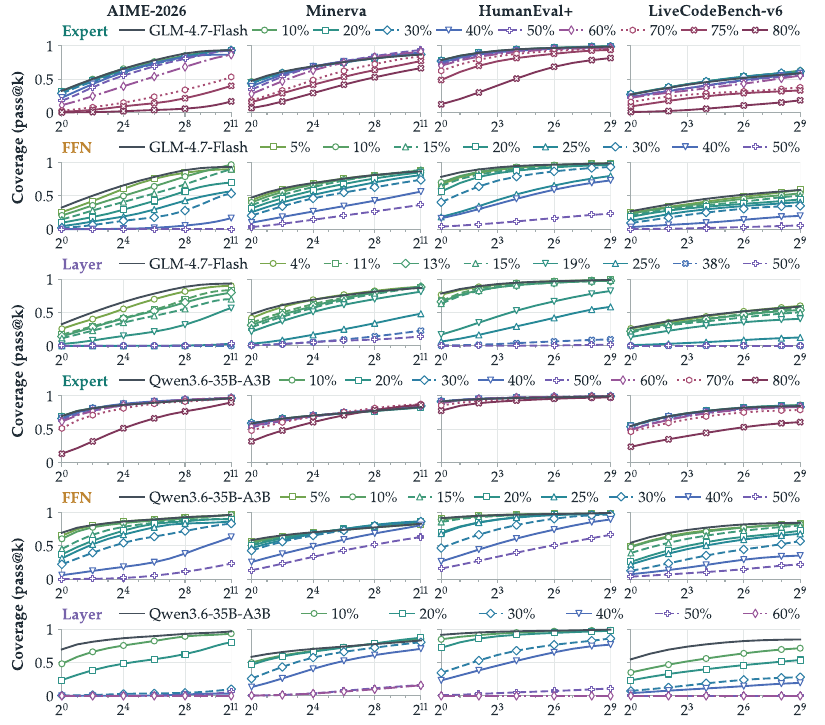}
\caption{\textbf{Pruned models can reach or exceed the original model's coverage.}
Pass@$k$ for Qwen3.6-35B-A3B and GLM-4.7-Flash across pruning axes, tasks and pruning ratios.}
\label{fig:teaser}
\end{figure}

\subsection{Measuring Reasoning Coverage}
\label{sec:evaluation}

For problem $q$, let $c_q$ of $n_q$ completions from model $m$ pass the
task's grader. For $1\le k\le\min_q n_q$, we use the pass@$k$ estimator
of~\citet{chen2021codex}
\begin{equation}
P_m(k)=\frac{1}{|\mathcal Q|}\sum_{q\in\mathcal Q}
\left[1-\frac{\binom{n_q-c_q}{k}}{\binom{n_q}{k}}\right].
\label{eq:passk}
\end{equation}
$P_m(1)$ is single-sample accuracy. At $k=n_q=n$, coverage is the observed
fraction of problems with at least one successful output. For smaller $k$,
the estimator averages coverage over uniformly sampled $k$-candidate subsets.
We use reasoning capacity to refer to the range of problems a model can
solve, and large-budget coverage as an empirical probe of that capacity.
The full curve shows how readily successful answers can be obtained as
the sampling budget increases, while matching problem identities
distinguishes retained, lost and newly observed successes.

Relative to the same backbone's original model, indexed by $0$, we report
\begin{equation}
E_m=\frac{P_m(1)}{P_0(1)},\qquad
R_{m,K}=\frac{P_m(K)}{P_0(K)}.
\label{eq:endpoints}
\end{equation}
$E_m$ measures accuracy retention and $R_{m,K}$ measures coverage retention.
Both can exceed one, and neither measures token efficiency. These
measurements depend on the prompts, decoder, output limit and grader.
They do not determine an unlimited-sampling ceiling: unobserved successes
may be rare rather than impossible. Grader acceptance also does not
establish the validity of a derivation.

Our default sampling settings are temperature $0.7$, top-$p=0.95$
\citep{holtzman2020nucleus}, decoding top-$k=20$ and an 8,192-token output cap.
For mathematics, the default is 2,048 responses per problem on all 30
AIME-2026\footnote{American Invitational Mathematics Examination, Mathematical
Association of America, 2026.
\url{https://maa.org/math-competitions/american-invitational-mathematics-examination-aime}}
and 272 Minerva problems~\citep{lewkowycz2022minerva}.
For code, the default comparison budget is $k=512$ on 164 HumanEval+
problems~\citep{liu2023evalplus} and a 175-problem LiveCodeBench-v6
subset~\citep{jain2025livecodebench}. Models within each comparison use
a common budget; budgets that differ from these defaults are specified
with the corresponding results.
Appendix~\ref{app:sampling} describes the sampling and grading procedures.

\section{Pruning Can Preserve or Expand Reasoning Coverage}
\label{sec:comparisons}

We first ask how much of the original model's coverage survives pruning.
Figure~\ref{fig:teaser} shows that repeated sampling can recover a broad
range of problems even after a substantial accuracy loss. To understand
what this preservation means, we begin with the amount of coverage
retained and then examine the identities of the solved problems.
These comparisons also reveal where preservation breaks down as the
pruning structure, task or source model changes.

\subsection{Retained Coverage and Newly Solved Problems}
\label{sec:survives}

\paragraph{Obs. 1. Substantial pruning can retain the original model's reasoning coverage.}
As the sampling budget grows, several pruned models reach or exceed
their original model's coverage (Figure~\ref{fig:teaser}). At 50\%
expert pruning, Qwen3.6-35B-A3B and GLM-4.7-Flash retain 100.6\% and
100.4\% of original coverage on average, even though they retain only
92.8\% and 75.8\% of original accuracy. These averages give equal
weight to the four task-wise retention ratios, using $K=2048$ for
mathematics and $K=512$ for code. The gap between accuracy and coverage
means that correct answers occur less frequently without a corresponding
contraction in the observed range of solvable problems. Expert pruning
provides the clearest examples of this preservation, while the FFN and
layer comparisons below show where additional sampling no longer recovers
the original model's coverage within the evaluated budgets.

\paragraph{Obs. 2. Pruned models can solve problems their original models did not solve.}
The coverage retained in Obs.~1 does not tell us whether pruning preserves
the same problems. We therefore compare which problems each model solves
at the same sampling budget (Appendix~\ref{app:solved-sets}). For
Qwen3.6-35B-A3B on AIME, 50\% expert pruning preserves all 29 problems
solved by the original model. At 80\%, however, it retains 26 of those
problems, loses three and solves one new problem. Increasing pruning can
thus change which problems are solved, even when many original successes
remain.

The Minerva comparison shows how this turnover can yield a net coverage
gain despite lower pass@1. At 80\% expert pruning of Qwen3.6-35B-A3B,
coverage increases from 226 to 235 of 272 problems because 23 newly
solved problems more than offset 14 lost successes. The larger solved
set therefore reflects a redistribution of successes, rather than
preservation of every original success followed by additional gains.
Grading differences qualify this comparison
(Appendix~\ref{app:sampling}), and the observed sets depend on the
sampling budget. A problem absent from one set may still admit rare
successes, so these changes do not determine which problems would remain
solvable under unlimited sampling (Section~\ref{sec:selection-scope}).

\subsection{Coverage Depends on the Pruning Axis and Task}
\label{sec:collapses}

\paragraph{Obs. 3. Expert pruning retains higher coverage than FFN pruning at similar model sizes.}
For Qwen3.6-35B-A3B and GLM-4.7-Flash, expert pruning retains
substantially more coverage than FFN pruning even when the remaining
model parameter counts are close (Table~\ref{tab:moe-coverage}). At 50\%
structural pruning, Qwen3.6-35B-A3B retains 18.54B parameters under
expert pruning and 18.49B under FFN pruning. The corresponding
GLM-4.7-Flash models retain 16.05B and 15.80B. Mean coverage retention
is approximately 100\% under expert pruning, compared with 48.4\%
and 18.8\%, respectively, under FFN pruning.

Similar total sizes do not mean that these models use the same amount
of computation per token. Expert pruning keeps routed top-$k$ fixed,
so it removes stored experts while leaving active parameter counts
nearly unchanged. FFN pruning instead narrows the modules used during
generation, reducing active parameters to 2.38B for Qwen3.6-35B-A3B
and 2.46B for GLM-4.7-Flash. The coverage advantage of expert pruning
thus accompanies greater preservation of active computation
(Figure~\ref{fig:severity} and Appendix~\ref{app:operators}).
Layer pruning goes further by removing attention as well as FFN
parameters. It retains only 9.4\% and 5.4\% of coverage at the nominal
50\% pruning ratio, although this comparison matches structural
ratios rather than total model size.

This advantage does not imply that expert pruning preserves coverage
equally well on every backbone. Gemma-4-26B-A4B-it loses coverage
earlier in its expert-pruning sweep (Appendix~\ref{app:backbone-sweeps}).
The distinction between removing width and removing depth also matters
when expert pruning is unavailable. In the dense Qwen3.5-4B model,
pruning 25\% of FFN width retains more coverage on both mathematics
tasks than pruning 25\% of layers, although both fall below the
original model (Appendix~\ref{app:dense-main-layout}).

\paragraph{Obs. 4. The same pruning operation can preserve coverage on one task and reduce it on another.}
\label{sec:task-agreement}
Even at a fixed pruning ratio within the same backbone, coverage can
change in opposite directions across tasks. Several
Qwen3.6-35B-A3B FFN and layer configurations exceed the original model's
Minerva coverage while losing coverage on AIME. Under severe pruning,
this divergence can leave successes on only one task.
Qwen3.6-35B-A3B layer-60\% retains nonzero Minerva coverage but has no observed
success on the other three tasks, and Gemma-4-26B-A4B-it layer pruning
exhibits a similar split (Appendix~\ref{app:four-task-sweeps}).
Retained coverage on one benchmark therefore need not extend to other tasks.

\begin{figure}[!t]
\centering
\includegraphics[width=\textwidth]{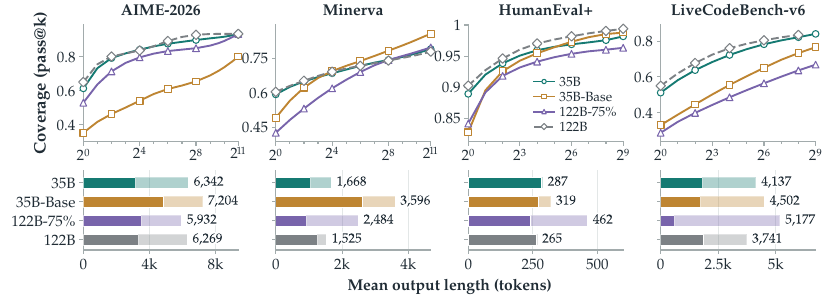}
\caption{\textbf{Coverage and output length across model sizes and training stages.}
35B and 35B-Base denote Qwen3.5-35B-A3B and Qwen3.5-35B-A3B-Base;
122B and 122B-75\% denote Qwen3.5-122B-A10B before and after 75\% expert pruning.
Coverage (top) and mean output tokens (bottom). Solid bars show means
over untruncated responses; pale extensions reach the means over all responses. LiveCodeBench-v6 curves
end at $k=256$ for 122B and $512$ for the other models.}
\label{fig:122b}
\end{figure}

\subsection{Does a Better Starting Model Give a Better Pruned Model?}
\label{sec:native-small}

\paragraph{Obs. 5. Starting from a larger or Instruct model does not guarantee higher coverage after pruning.}
A model may preserve much of its original coverage after pruning yet
still underperform a smaller original model. To examine whether a better
starting model yields a better pruned model, we extend accuracy-based
comparisons~\citep{jaiswal2024llmkick} to Base and Instruct releases, dense models matched by FFN width,
and a larger MoE pruned to a smaller size.

The Base/Instruct ordering changes with both task and pruning axis.
Before pruning, Qwen3.5-35B-A3B-Base leads Instruct on Minerva but trails it on
AIME. Its Minerva lead survives 50\% expert pruning and reverses
under 50\% FFN pruning, using the same grading policy
(Appendix~\ref{app:inheritance}). Neither the original ordering nor
the accuracy change alone predicts coverage after pruning. These
results depend on which release is pruned, but do not isolate the effects
of individual post-training stages.

Comparisons across model sizes require a further distinction between
the size of the source and the size remaining after pruning. Lightly
FFN-pruned Qwen3.5-9B exceeds original Qwen3.5-2B on AIME and
LiveCodeBench-v6, but not on Minerva, while still being much larger.
Matching the FFN width gives a different comparison. At 25\% FFN
pruning, the pruned 9B model has the same FFN width as Qwen3.5-4B,
yet trails the original 4B model in both accuracy and coverage on all
four tasks (Appendix~\ref{app:inheritance}). A larger source model
therefore does not guarantee better reasoning performance at a
comparable retained width.

The larger MoE illustrates why preserving source coverage and outperforming
a smaller original model are separate comparisons. At 75\% expert pruning,
Qwen3.5-122B-A10B retains 28/30 AIME problems even as pass@1 falls from
65.0\% to 53.0\%. This substantial retention only matches original
Qwen3.5-35B-A3B on AIME, however. The pruned model slightly exceeds it
on Minerva but trails it on both code tasks (Figure~\ref{fig:122b}; Appendix~\ref{app:inheritance}).
Across these comparisons, retaining a larger source's coverage gives no
consistent advantage over a smaller original model. Because neither total
parameters nor training compute are matched, these comparisons do not
isolate the effect of starting model size.

\section{Pruning Changes CoT Diversity and Generation Reliability}
\label{sec:process}

The previous section shows that a model can retain broad coverage after
pruning. Whether that coverage remains useful depends on how readily
the model generates correct answers. More attempts increase the cost
of reaching a target, while changes in response length can amplify or
offset the increase (Figure~\ref{fig:length};
Appendix~\ref{app:fixed-targets}). We first quantify this cost, then
examine termination, repetition within responses and variation across
responses to characterize how generation changes after pruning.

\subsection{Reaching the Same Coverage Can Require Many More Tokens}
\label{sec:token-use}

\paragraph{Obs. 6. Severe pruning can sharply increase the sampling cost of reaching the same coverage.}
To determine the cost of recovering coverage, we count both the attempts
required to reach a shared target and the tokens generated per attempt.
We estimate mean generated tokens per problem as $k_m^*\bar t_m$, where
$k_m^*$ is the smallest number of candidates attaining the target and
$\bar t_m$ is mean output length. On Qwen3.6-35B-A3B AIME, 50\% expert
pruning reaches 90\% of the original model's coverage in 22 candidates,
compared with 20 for the original model, and requires 1.12 times as many
generated tokens. At 80\% expert pruning, reaching the same target
requires 1,319 candidates and 59.63 times the tokens. Mean response
length actually falls from 4,923.5 to 4,451.4 tokens, but that saving per
attempt is overwhelmed by the number of attempts needed
(Appendix~\ref{app:fixed-targets}). Thus, moderate pruning can leave
generated-token cost close to the original model's, whereas severe
pruning can raise it sharply even with shorter responses. This measure
excludes prompt processing and verification and does not measure hardware
speedup (Appendix~\ref{app:cost}).

\subsection{Severe Pruning Changes How Attempts Fail}
\label{sec:stability}

\paragraph{Obs. 7. Lower truncation does not imply more reliable generation.}
On AIME, severe layer pruning produces many normally terminated but
incorrect responses in both Qwen3.6-35B-A3B and GLM-4.7-Flash.
As pruning increases, lower truncation can therefore reflect a change in
how attempts fail rather than a recovery in correctness
(Appendix~\ref{sec:termination-reliability}).

Even problems answered consistently by the original model can fail
frequently without reaching the length cap. Consider LiveCodeBench-v6
problems that Qwen3.6-35B-A3B answers correctly in every original sample.
For the half with longer original responses, 30\% FFN pruning yields
29.94\% pass@1 but only
2.08\% length truncation. At least 67.98\% of all outputs are therefore
incorrect and not length-truncated, while all 14 problems still have a
successful sample. The model thus retains a successful response for every
problem in this group, yet usually fails before reaching the output cap
(Appendix~\ref{app:within-task-cost}).

The distinction between stopping and succeeding also changes how we
interpret response length. If a model increasingly terminates with an
incorrect answer, a shorter response need not indicate more efficient
reasoning. On Qwen3.6-35B-A3B AIME, raising layer pruning from 30\%
to 60\% reduces mean length from 7,806 to 1,564 tokens while
pass@512 falls from 5.6\% to zero. The apparent saving in output
length therefore accompanies a complete loss of observed successes
at that budget.

\begin{figure}[!t]
\centering
\includegraphics[width=\textwidth]{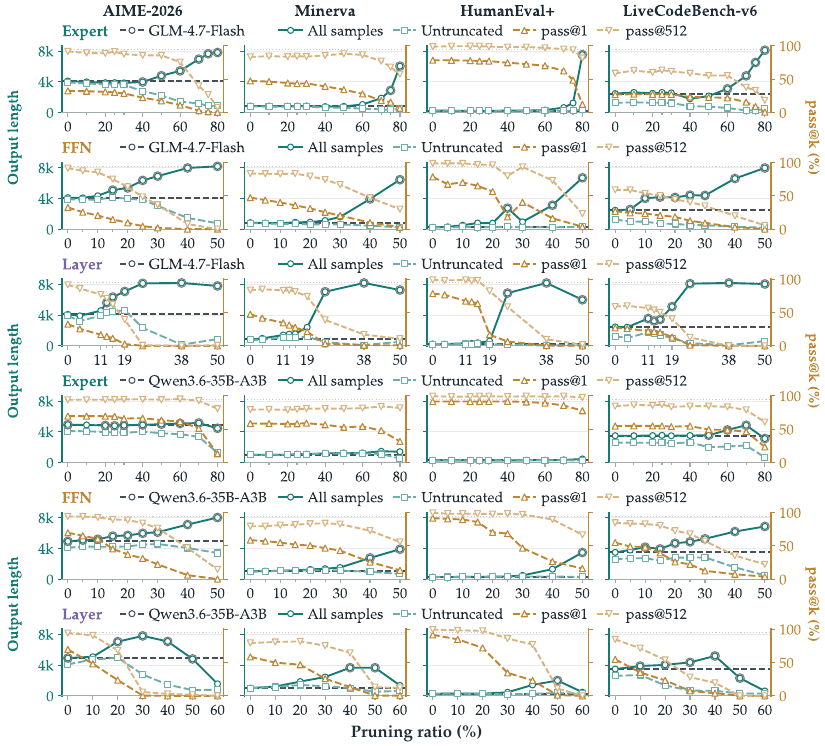}
\caption{\textbf{Output length, accuracy and coverage across pruning ratios.}
Results are shown for Qwen3.6-35B-A3B and GLM-4.7-Flash.
Left axes show mean output tokens (thousands), right axes pass@1/pass@512.
Grey rings mark $\geq20\%$ truncation. Figure~\ref{fig:gemma-length}
shows the corresponding results for Gemma-4-26B-A4B-it.}
\label{fig:length}
\end{figure}

To distinguish shorter generation from changes in truncation,
Figure~\ref{fig:length} separates the mean over all outputs from the
mean over untruncated outputs. The latter describes only responses
that finish before the cap, so the gap between the means also depends
on how often the remaining responses reach the cap and how long they
are. A complementary test is to allow generation to continue longer.
On Qwen3.6-35B-A3B expert-30\%, raising the cap from 8k to 16k
improves AIME pass@1 by 7.70 points and coverage at $k=2048$ from
29 to 30 problems. Layer-30\% remains near 0.37\% pass@1, although
coverage at the same budget rises from three to five problems
(Appendix~\ref{sec:decoding}). The larger cap therefore recovers
some successes, but its benefit depends on the pruning structure
and does not uniformly restore reliability.

\subsection{Greater CoT Diversity Can Coexist with Repetition}
\label{sec:trace-analysis}

\begin{figure}[!t]
\centering
\includegraphics[width=\textwidth]{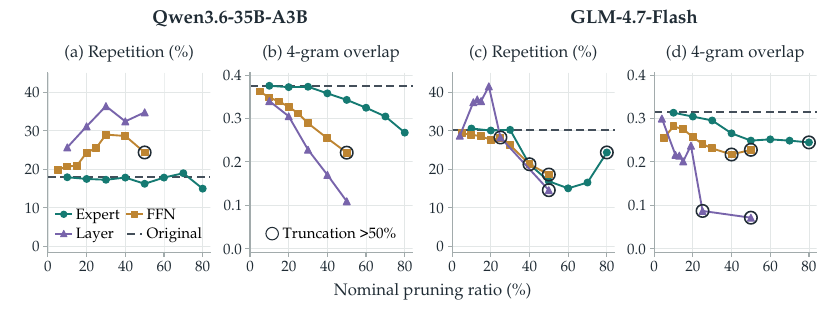}
\caption{\textbf{Repetition within responses and overlap across candidates.}
Equal-weight averages over AIME-2026, Minerva, HumanEval+ and
LiveCodeBench-v6 use 2,048 candidates per math problem and 512 per code
problem. Panels (a,b) show Qwen3.6-35B-A3B; (c,d) show GLM-4.7-Flash.
Within each task, (a,c) give the percentage of normally terminated outputs
containing an 8-gram at least three times; (b,d) average 4-gram overlap
equally across eligible problems. Both metrics include incorrect outputs.
Rings mark mean truncation above 50\%.}
\label{fig:process}
\end{figure}

\paragraph{Obs. 8. Pruning can increase CoT diversity across samples even as repetition grows within responses.}
A response may repeat itself without resembling the other responses
sampled for the same problem. We examine both properties among normally
terminated outputs, including incorrect responses. Repetition is the
fraction containing an 8-gram at least three times; diversity is measured
by symmetric clipped 4-gram overlap between responses, with lower overlap
indicating greater textual diversity. Averaged equally across the four
tasks, 19\% layer pruning on GLM-4.7-Flash raises repetition from 30.2\%
to 41.5\% while reducing overlap from 0.316 to 0.237
(Figure~\ref{fig:process}). Thus, more responses repeat text internally
even as the sampled outputs share less text with one another.

Greater diversity need not always come with more repetition. In the
same four-task average, GLM-4.7-Flash expert-50\% reduces overlap to
0.249 while also lowering repetition to 16.9\%. Together with the
layer-pruning result, this shows why variation across responses and
repetition within them must be examined separately. Their behavior
also changes along the pruning sweep. GLM-4.7-Flash AIME expert-80\%
has higher overlap than the original model, and intermediate ratios
can reverse the direction as well. Because these metrics use normally
terminated outputs, high truncation limits how well they describe the
full response pool. They measure diversity in generated CoT text,
including incorrect responses, rather than the number of valid solution
strategies. Even a correct final answer can accompany an invalid
derivation (Appendix~\ref{app:reasoning-counterexample}).

\section{Original Accuracy Can Conceal Pruning Vulnerability}
\label{sec:task-demands}

The cost and failure patterns above show that preserved coverage can
hide substantial deterioration. Can we identify vulnerable problems
from the original model's behavior, before applying pruning? Longer
responses and lower original accuracy both suggest greater task demands,
but they often occur together. We first examine their association with
coverage loss across tasks, then compare problems with the same or similar
original success frequencies to ask what length tells us beyond accuracy.

\subsection{Pruning Sensitivity Depends on Task Demands}
\label{sec:cross-task-demands}

\paragraph{Obs. 9. More demanding reasoning tasks may be more vulnerable to pruning.}
The code benchmarks in Figure~\ref{fig:length} provide a clear comparison.
For both Qwen3.6-35B-A3B and GLM-4.7-Flash, original responses are longer
and single-sample accuracy is lower on LiveCodeBench-v6 than on HumanEval+.
For Qwen3.6-35B-A3B, the original model averages 3,489 versus 270 tokens,
with pass@1 of 54.8\% versus 91.5\%. As FFN pruning increases from 10\%
to 30\%, LiveCodeBench-v6 retains 98.0\% and then 67.6\% of its original
pass@512, while HumanEval+ retention changes only from 98.8\% to 97.5\%.
LiveCodeBench-v6 responses also lengthen to 5,261 tokens at 30\% pruning,
so the coverage loss is accompanied by more generation per attempt.
GLM-4.7-Flash shows the same pattern, retaining 59.2\% of original
coverage on LiveCodeBench-v6 versus 94.4\% on HumanEval+ at 30\% FFN
pruning. This sensitivity varies with pruning structure, with
Qwen3.6-35B-A3B expert pruning providing exceptions to the cross-task
trend (Appendix~\ref{app:task-length}).

\subsection{Consistent Original Success Does Not Ensure Pruning Robustness}
\label{sec:within-task}

\paragraph{Obs. 10. Among consistently solved code problems, the group with longer original responses loses more accuracy under FFN pruning.}
The cross-task comparison leaves open whether longer responses add
information beyond lower original accuracy. To separate these two
properties, we restrict attention to problems answered correctly in
every original sample and split them into equal groups by original
mean response length. On Qwen3.6-35B-A3B LiveCodeBench-v6, both groups
therefore begin with 100\% observed accuracy. Nevertheless, 30\% FFN
pruning reduces pass@1 to 65.0\% for the shorter-response half and
29.9\% for the longer-response half, with 14 problems in each. Equal
success before pruning thus conceals a substantial difference in
robustness afterward.

The longer-response group also loses more accuracy on HumanEval+ for
both backbones. This direction holds at each tested FFN pruning ratio
from 5\% to 50\% among code problems answered correctly in every
original sample. To move beyond this subset, we group problems within two-percentage-point bands
of original success frequency and compare the shorter and longer halves
within each band. After pooling the bands and measuring each group's
loss from its own original accuracy, the longer group has a larger
mean decline on both code tasks and both backbones at each tested
ratio in the same range. Wider bands broadly preserve the code pattern,
whereas mathematics is more sensitive to grouping
(Appendix~\ref{app:within-task}). Original response length thus adds
information beyond observed accuracy. Because full responses include
code and explanations, this association may reflect several task demands;
it does not isolate CoT length as a cause of pruning sensitivity.

This vulnerability matters even when repeated sampling eventually
recovers every problem, because lower accuracy and longer responses
can both increase the cost of obtaining correct answers. To examine
their combined effect, we take a separate group consisting of the
longer-response half of GLM-4.7-Flash LiveCodeBench-v6 problems with
at least 90\% original success frequency. At 30\% FFN pruning, all
14 problems remain covered at $k=512$, but pass@1 falls from 96.3\%
to 29.9\% and mean output length grows from 436 to 1,873 tokens.
Together, less frequent success and longer attempts raise generated
tokens per correct sample 13.84-fold. This ratio divides total sampled
tokens by the number of correct samples, so it measures the cost of
the successes in the pool rather than the cost of reaching a fixed
coverage target or obtaining a first success on each problem
(Appendix~\ref{app:within-task-cost}).

\section{Pruning Method Gains Can Reflect Sampling Efficiency}
\label{sec:method-comparison}

With the pruning structure and ratio fixed, selection methods may still
differ in how often they solve a problem and which problems they solve.
A pass@1 gain alone does not distinguish these effects. We compare the coverage curves
and solved sets of RAZOR (Section~\ref{sec:setup}), expert activation
norm (EAN) and routing frequency (FREQ) at fixed pruning ratios on
Qwen3.6-35B-A3B (Appendix~\ref{app:selection-controls}), then examine
the token cost of reaching the same target.

\begin{figure}[!t]
\centering
\includegraphics[width=\textwidth]{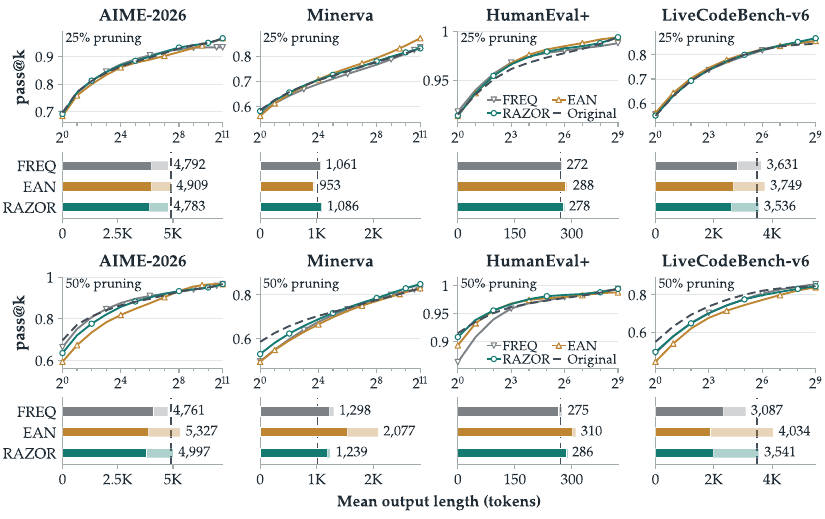}
\caption{\textbf{Qwen3.6-35B-A3B expert selection at 25\% and 50\% pruning.}
Each block shows pass@$k$ above and mean output tokens below, with shared
token scales across pruning ratios. Solid bars show means over untruncated
responses; pale extensions reach the means over all responses. Mathematics uses $n=2048$
and code $n=512$.}
\label{fig:expert-strategy}
\end{figure}

\subsection{The Same Problems Can Be Solved with Fewer Attempts}
\label{sec:selection-efficiency}

\paragraph{Obs. 11. Higher single-sample accuracy can reflect more frequent success on the same problems.}
On Qwen3.6-35B-A3B at 50\% expert
pruning, RAZOR has higher AIME pass@1 than EAN, at 63.53\% versus
59.38\%, yet both solve the same 29 problems at $k=2048$.
FREQ also solves exactly this set. To reach the shared 87.0\%
coverage target, corresponding to 90\% of the original endpoint,
RAZOR, EAN and FREQ need 22, 58 and 14 candidates per problem,
respectively (Figure~\ref{fig:expert-strategy}). The methods therefore
differ in sampling efficiency even when their observed solved sets coincide.

The code tasks show a similar contrast between accuracy and coverage,
although their endpoint coverage does not agree exactly. Across
the three criteria, pass@1 spans 4.48 points on HumanEval+ and 5.26
on LiveCodeBench-v6, whereas coverage at the largest common budget
differs by one problem and 1.71 points, respectively
(Appendix~\ref{app:selection-controls}). The smaller coverage gaps
suggest that much of the accuracy difference concerns how readily
successes are sampled. They do not by themselves show that the methods
solve the same problems, since similar coverage counts can conceal
different gains and losses.

\subsection{Output Length Also Affects Selection Efficiency}
\label{sec:selection-work}

Fewer attempts need not translate into proportionally fewer tokens.
We therefore apply the fixed-target token measure from
Section~\ref{sec:token-use} to the selection methods.
For Qwen3.6-35B-A3B at 50\% expert pruning, RAZOR reaches the shared
87.0\% AIME target with an estimated 109,923 generated tokens per problem,
compared with 308,985 for EAN, reducing generated-token use by 64.4\% at the
same target. The length differences in Figure~\ref{fig:expert-strategy}
show why accuracy alone cannot determine the token cost of sampling.
These estimates allocate the same number of candidates to every problem
and use the mean length over all responses, with no stopping after success
(Appendix~\ref{app:selection-controls}).

\subsection{Coverage Convergence Differs Across Tasks}
\label{sec:selection-scope}

\paragraph{Obs. 12. Expert-selection methods agree on the AIME solved set, while coverage differences persist on Minerva.}
The agreement on AIME raises the question of whether expert-selection
differences generally disappear with more sampling. At $k=2048$, the
methods solve exactly the same AIME problems, providing a clear example
of improved sampling efficiency without broader observed coverage.
Minerva does not show the same agreement. Coverage still differs across
selection methods and pruning ratios at that budget
(Figures~\ref{fig:teaser} and \ref{fig:expert-strategy}, with grading
conventions in Appendix~\ref{app:sampling}). The AIME result therefore
explains how a pass@1 gain can arise without extending the solved set,
but does not establish that every method comparison has this form.
Nor do the remaining Minerva gaps establish a permanent separation.
Both agreement and disagreement at finite budgets leave open whether
the methods would eventually solve the same problems with unlimited sampling.

\section{Combining Pruning Structures Need Not Improve the Size--Capacity Trade-off}
\label{sec:mixed}

Given the coverage retained by expert pruning, we ask whether distributing
compression across structures improves coverage at a given model size.
We combine Expert, FFN and Layer pruning in two Qwen3.6-35B-A3B models,
denoted Mixed-A and Mixed-B. Both use 50\% expert pruning and
10\% layer pruning, with approximately 20\% and 30\% FFN pruning, respectively.
FFN pruning reduces the intermediate width of both routed and shared
experts. Both models use RAZOR for expert selection,
omit the multi-token prediction module and receive no recovery training,
giving total sizes of 13.82B and 12.45B under the parameter
accounting in Table~\ref{tab:moe-coverage}. Table~\ref{tab:mixed-results}
compares them with expert-only RAZOR at 70\%, 75\% and 80\% pruning
(Appendix~\ref{app:mixed-design}).

\paragraph{Obs. 13. A tested pruning combination loses coverage at nearly the same total model size.}
Mixed-B has a similar total size to expert-70\%, with 12.45B versus
12.12B parameters, but uses fewer active parameters per token,
2.39B versus 2.93B. Both counts exclude vision and MTP modules,
as in Table~\ref{tab:moe-coverage}. Lower active parameter use accompanies
lower accuracy and coverage on AIME and both code tasks
at the common task budgets. The reported
Minerva scores have the same ordering, subject to grading differences
(Appendix~\ref{app:sampling}).

\begin{figure}[!t]
\centering
\includegraphics[width=\linewidth]{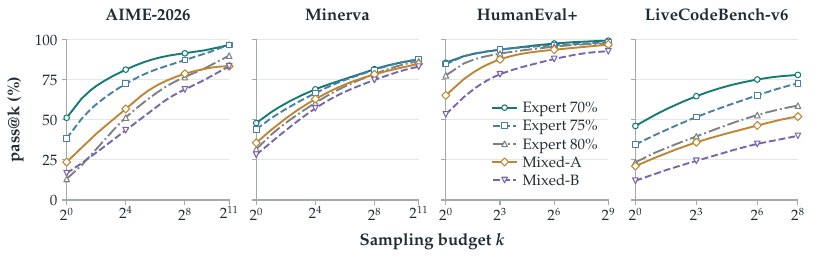}
\caption{\textbf{Coverage across sampling budgets for Qwen3.6-35B-A3B.}
Expert-only and mixed configurations are defined in Table~\ref{tab:mixed-parameters}.
The base-2 axis ends at the largest common budget: $k=2048$ for mathematics,
$512$ for HumanEval+ and $256$ for LiveCodeBench-v6.
Minerva graders differ (Appendix~\ref{app:sampling}).}
\label{fig:mixed-budget}
\end{figure}

The comparison also depends on how much sampling is allowed. Mixed-A
leads expert-80\% on AIME at $k=64$, but this ordering reverses as
the budget grows (Figure~\ref{fig:mixed-budget}). At $k=2048$, both
mixed models cover 25 problems, compared with 29, 29 and 27 for
expert-only pruning at 70\%, 75\% and 80\%. These models differ in
size, so the comparison does not isolate the allocation of pruning.
It does show why a low-budget advantage is insufficient to conclude
that the retained structures support a broader range of problems.

Even within the mixed configurations, equal coverage does not mean
equivalent behavior. Mixed-B has lower AIME pass@1 than Mixed-A,
16.7\% versus 23.6\%, while both cover 25 problems at $k=2048$.
Each also solves one problem the other misses, so the same coverage
count conceals differences in both success frequency and problem
identity. On the other tasks, Mixed-B has lower coverage at the common
budgets. Although it has a narrower FFN, the two variants also retain
different experts, preventing attribution of these differences to
FFN pruning alone.

These comparisons distinguish total size from computation per token.
Expert pruning preserves routed top-$k$, whereas FFN and layer pruning
reduce active parameters. The models also differ in whether they retain
the MTP module and in pruning order, so the performance gap cannot be
attributed solely to how pruning is distributed across structures
(Appendix~\ref{app:mixed-design}). Fewer active parameters alone do not
establish faster inference.

\section{Discussion}
\label{sec:discussion}

Across the comparisons, some problems remain easy to solve, others
become rare successes, and some exchange places in the solved set.
To understand how pruning redistributes success, we first ask whether
these changes can be explained by making every problem harder by the
same factor. We test this explanation with a reference matched to the
mean accuracy loss, then consider whether gains on some problems could
reflect domain specialization.

\subsection{Does Pruning Make All Problems Uniformly Harder?}
\label{sec:uniform-discussion}

\paragraph{Obs. 14. A uniform drop in success probability does not explain all coverage losses.}
The uniform-loss reference independently retains sampled successes from
the original model at a common rate chosen to match the pruned model's
observed pass@1 in expectation (Appendix~\ref{app:thinning}). It asks how
much coverage would remain if pruning reduced success uniformly across
problems.

Comparing observed coverage with this reference tests what the
matched mean accuracy leaves unexplained. For Qwen3.6-35B-A3B FFN-40\%
on AIME, observed coverage is 63.3\% against the reference's
92.1\%, a $-28.7$-point residual with a 95\% interval of
[$-45.2$, $-13.4$] from paired bootstrap resampling of problems.
Its LiveCodeBench-v6 residual is $-46.9$ points [$-53.8$, $-40.0$],
with equally sized 512-candidate pools. The intervals condition on
the generated responses, and the discrepancies show that matching
the average accuracy loss does not reproduce the coverage curve.
Pruning must therefore be examined at the problem level as well.
Changes in difficulty ordering and solved-set membership provide
complementary views of this uneven effect
(Appendices~\ref{app:rank} and~\ref{app:solved-sets}).

\subsection{Could Coverage Gains Reflect Domain Specialization?}

Domain specialization is one possible explanation for gains on some
problems alongside losses on others. Pruning could preferentially retain
computation useful for a subset of topics and alter the distribution of
successful responses. The gains on some tasks and changes in which
problems are solved motivate this hypothesis, but do not establish it. Aggregate pass@$k$
does not reveal whether newly solved problems cluster by topic, and
survival on one task after losses elsewhere does not itself demonstrate
improvement in that domain. Establishing specialization requires
gains and losses within topics under the same evaluation conditions.

\section{Related Work}
\label{sec:related}

\paragraph{Structured pruning.}
Structured pruning methods differ in the computation they remove and
the compensation used to preserve performance. Width methods select
coupled parameters, compensate for their removal or reduce dimensions
\citep{ma2023llmpruner,an2024flap,van2024llmsurgeon,ashkboos2024slicegpt},
whereas depth methods remove or collapse blocks
\citep{men2024shortgpt,gromov2024unreasonable,song2024sleb,yang2024laco}.
Training-based pruning can optimize several structures jointly
\citep{xia2024shearedllama}. MoE models offer another choice because
their experts can be pruned~\citep{lu2024expertsparsity} or merged
\citep{he2023mergemoe,ream2026}, with FFE examining the roles of
selection, pruning and recovery~\citep{ffe2025}. These choices produce
different compressed models, making it useful to ask not only how
accurate each model remains, but which problems it can still solve
and how much sampling those successes require.

\paragraph{Reasoning after compression and training.}
The performance of a compressed model also depends on whether pruning
is followed by recovery training. Studies of layer removal and sparse
allocation examine how compression affects test-time scaling
\citep{wang2026layerpruning,monjur2026revisiting}. ReCal connects
pruning to subsequent recovery by reweighting calibration to preserve
teacher-supported predictions, improving reasoning performance after
offline distillation followed by on-policy distillation (OPD)
\citep{jiang2026recal}. Our comparisons examine the models before
recovery training. This allows us to characterize the coverage and
generation behavior left by expert, FFN and layer pruning without
combining those changes with the effects of further training.

\paragraph{Measuring and using sampled solutions.}
Repeated-sampling analyses of code and reasoning
\citep{chen2021codex,brown2024monkeys,yue2025rlvr} distinguish
single-attempt accuracy from the range of problems solved across
multiple attempts. A recent survey of OPD likewise emphasizes that
accuracy gains need not imply broader solution coverage
\citep{song2026opdsurvey}. We apply this distinction to pruning by
examining which problems remain solvable and how readily successes
can be sampled. Using those successes in practice also depends on
how candidates are generated and selected. Chain-of-thought
prompting~\citep{wei2022cot} and inference allocation
\citep{snell2025testtime,muennighoff2025s1} shape their generation,
while self-consistency~\citep{wang2023selfconsistency} and
verifiers~\citep{lightman2024verify} help select an answer. Candidate quality also
requires attention to behavior that an aggregate score can hide,
as emphasized by broad capability, behavioral and joint efficiency
evaluations~\citep{dutta2024accuracy,vonrad2026unicomp}.
For pruning in particular, \citet{jaiswal2024llmkick} report
downstream degradation and more repetitive generation even when
perplexity changes little. This motivates examining generated
responses alongside the coverage and accuracy they achieve.

\section{Conclusion}
\label{sec:conclusion}

Structured pruning can preserve a broad range of solvable problems while
making successful answers much harder to obtain. Lower accuracy
therefore need not mean that reasoning coverage has narrowed, just as
higher pass@1 need not mean that a pruning method preserves a broader
range of solvable problems. The distinction becomes consequential under
severe pruning, where increasingly rare successes can drive up total
token use despite shorter individual responses. Retained coverage must
therefore be evaluated alongside the cost of recovering it.

Understanding this trade-off requires looking at what pruning removes
and where the resulting losses occur. At similar total model sizes,
expert pruning retains more coverage alongside more active parameters
than FFN pruning or the tested combination of pruning structures.
Stored model size alone therefore does not explain the retained
reasoning performance. Nor does original accuracy fully describe
which problems are vulnerable, since original response length can
reveal differences hidden by similar success frequencies, with an
association that varies across tasks. Evaluating pruning through
matched-budget coverage, sampling cost and total and active parameter
counts makes these distinctions visible, while finite-budget
observations leave ultimate reasoning capacity unresolved.

\phantomsection
\label{page:mainend}

\section*{AI use statement}
We used generative AI tools for language polishing, literature search,
coding assistance and refinement of the presentation and analysis of
experimental results. The authors are responsible for the final text, claims
and artifacts.

\section*{Ethics statement}
Compression does not remove the risks of unreliable outputs. Aggregate
coverage can conceal failures on individual problems, and grader-accepted
answers can contain invalid derivations. Our problem-level and generation
analyses make these failures visible. Resource benefits also require
careful measurement: structural pruning ratios, parameter counts and
generated-token use capture different costs, while latency and energy
require direct evaluation.

\section*{Reproducibility statement}
Appendix~\ref{app:protocol} specifies model releases, pruning methods,
generation settings, task subsets, sample counts, length caps and grading.
The appendix contents locate the supporting results, estimators and
sensitivity analyses.
\bibliography{references,references_added}
\bibliographystyle{colm2026_conference}
\clearpage

\appendix

\setcounter{secnumdepth}{3}

\section*{Appendix contents}
\pdfbookmark[1]{Appendix contents}{appendix-contents}
\begingroup
\setcounter{tocdepth}{2}
\hypersetup{linktoc=all}
\makeatletter
\@starttoc{atoc}
\makeatother
\endgroup
\bigskip

\makeatletter
\let\appendix@addcontentsline\addcontentsline
\renewcommand{\addcontentsline}[3]{%
  \appendix@addcontentsline{#1}{#2}{#3}%
  \def\appendix@extension{#1}%
  \def\appendix@toc{toc}%
  \ifx\appendix@extension\appendix@toc
    \def\appendix@level{#2}%
    \def\appendix@section{section}%
    \def\appendix@subsection{subsection}%
    \ifx\appendix@level\appendix@section
      \addtocontents{atoc}{\protect\contentsline{#2}{#3}{\thepage}{\@currentHref}\protected@file@percent}%
    \else\ifx\appendix@level\appendix@subsection
      \addtocontents{atoc}{\protect\contentsline{#2}{#3}{\thepage}{\@currentHref}\protected@file@percent}%
    \fi\fi
  \fi
}
\makeatother

\FloatBarrier
\section{Experimental Design and Measurement}
\label{app:protocol}

\subsection{Models, architectures and comparison design}
\label{app:coverage}
\label{app:architectures}

To examine how coverage depends on pruning structure and backbone, we compare
expert, FFN and layer pruning on Qwen3.6-35B-A3B, GLM-4.7-Flash and
Gemma-4-26B-A4B-it (Table~\ref{tab:moe-coverage}).
Each model generates 2,048 responses per problem on all 30 AIME-2026
and 272 Minerva problems. HumanEval+ and LiveCodeBench-v6 comparisons
use $K=512$ (Appendix~\ref{app:code-endpoints}).
The Qwen3.6-35B-A3B AIME results at 25\%, 30\% and 40\% expert pruning
are outcome-selected and are excluded from statistical certification.

\begin{table}[!t]
\centering
\small
\begingroup
\fontsize{8}{9}\selectfont
\renewcommand{\arraystretch}{1.05}
\setlength{\tabcolsep}{2pt}
\begin{tabular*}{0.92\textwidth}{@{\extracolsep{\fill}}crr@{\hspace{10pt}}crr@{\hspace{10pt}}crr@{}}
\toprule
\multicolumn{3}{c}{\textbf{Qwen3.6-35B-A3B}} & \multicolumn{3}{c}{\textbf{GLM-4.7-Flash}} & \multicolumn{3}{c}{\textbf{Gemma-4-26B-A4B-it}}\\
\cmidrule(lr){1-3}\cmidrule(lr){4-6}\cmidrule(lr){7-9}
\textbf{Ratio (\%)} & \textbf{Total (B)} & \textbf{Active (B)} & \textbf{Ratio (\%)} & \textbf{Total (B)} & \textbf{Active (B)} & \textbf{Ratio (\%)} & \textbf{Total (B)} & \textbf{Active (B)}\\
\midrule
\multicolumn{9}{@{}l}{\textbf{Original models}}\\
0 & 34.66 & 2.95 & 0 & 29.94 & 3.58 & 0 & 25.23 & 3.82\\
\midrule
\multicolumn{9}{@{}l}{\textbf{(a) Expert pruning}}\\
10 & 31.39 & 2.94 & 10 & 27.34 & 3.58 & 10 & 22.91 & 3.82\\
20 & 28.24 & 2.94 & 20 & 24.30 & 3.58 & 20 & 20.59 & 3.82\\
25 & 26.60 & 2.94 & 25 & 23.00 & 3.58 & 25 & 19.52 & 3.82\\
30 & 24.97 & 2.94 & 30 & 21.69 & 3.58 & 30 & 18.45 & 3.82\\
40 & 21.82 & 2.94 & 40 & 18.65 & 3.58 & 40 & 16.13 & 3.82\\
50 & 18.54 & 2.94 & 50 & 16.05 & 3.58 & 50 & 13.81 & 3.82\\
60 & 15.27 & 2.93 & 60 & 13.44 & 3.58 & 60 & 11.49 & 3.82\\
70 & 12.12 & 2.93 & 70 & 10.40 & 3.58 & 70 & 9.17 & 3.81\\
75 & 10.49 & 2.93 & 75 & 9.10 & 3.58 & 75 & 8.10 & 3.81\\
80 & 8.85 & 2.93 & 80 & 7.80 & 3.57 & 80 & 7.03 & 3.81\\
\midrule
\multicolumn{9}{@{}l}{\textbf{(b) FFN pruning}}\\
5 & 33.14 & 2.89 & 5 & 28.47 & 3.46 & 5 & 24.17 & 3.73\\
10 & 31.63 & 2.84 & 10 & 27.14 & 3.36 & 10 & 22.84 & 3.62\\
15 & 29.61 & 2.77 & 15 & 25.67 & 3.24 & 15 & 21.78 & 3.53\\
20 & 28.09 & 2.72 & 20 & 24.35 & 3.14 & 20 & 20.45 & 3.42\\
25 & 26.58 & 2.66 & 25 & 22.87 & 3.02 & 25 & 19.39 & 3.33\\
30 & 25.06 & 2.61 & 30 & 21.40 & 2.91 & 30 & 18.33 & 3.24\\
40 & 21.52 & 2.49 & 40 & 18.60 & 2.68 & 40 & 15.93 & 3.04\\
50 & 18.49 & 2.38 & 50 & 15.80 & 2.46 & 50 & 13.55 & 2.84\\
\midrule
\multicolumn{9}{@{}l}{\textbf{(c) Layer pruning}}\\
10 & 31.30 & 2.70 & 4 & 28.67 & 3.44 & 20 & 20.33 & 3.21\\
20 & 27.93 & 2.46 & 11 & 26.77 & 3.23 & 40 & 15.44 & 2.59\\
30 & 24.57 & 2.22 & 13 & 26.13 & 3.17 &  &  & \\
40 & 21.20 & 1.97 & 15 & 25.50 & 3.10 &  &  & \\
50 & 17.84 & 1.73 & 19 & 24.23 & 2.96 &  &  & \\
60 & 14.47 & 1.48 & 25 & 22.32 & 2.75 &  &  & \\
 &  &  & 38 & 18.51 & 2.34 &  &  & \\
 &  &  & 50 & 14.70 & 1.92 &  &  & \\
\bottomrule
\end{tabular*}
\endgroup
\caption{\textbf{Remaining total and active parameters at each pruning ratio.}
B denotes billions. Ratios specify the fraction of experts, FFN width or
layers pruned; 0\% denotes the original model. Counts follow the
parameter definitions in Appendix~\ref{app:operators}.}
\label{tab:moe-coverage}
\end{table}

For Gemma-4-26B-A4B-it, layer pruning removes complete six-layer units
from thirty layers. The evaluated range ends at 40\%, where output is
already degenerate (Figure~\ref{fig:fourtask-gemma}).

Dense models test how architecture affects accuracy and coverage retention
under FFN and layer pruning. We evaluate
Qwen3.5-9B\footnote{\url{https://huggingface.co/Qwen/Qwen3.5-9B}},
Qwen3.5-4B\footnote{\url{https://huggingface.co/Qwen/Qwen3.5-4B}},
Qwen3.5-2B\footnote{\url{https://huggingface.co/Qwen/Qwen3.5-2B}} and
Llama-3.1-8B-Instruct~\citep{llamateam2024llama3}, pairing each with
its original model. The original Qwen3.5-9B, Qwen3.5-4B and Qwen3.5-2B
models are not matched in total parameter count to the pruned models.
Tail-drop pruning provides a control for calibrated depth selection.

At fixed pruning ratios on Qwen3.6-35B-A3B, we compare RAZOR,
expert activation norm (EAN) and routing frequency to distinguish
sampling efficiency from reasoning capacity.

We also examine whether the capabilities retained after pruning depend
on the original model's training stage or size. We compare the
Base\footnote{\url{https://huggingface.co/Qwen/Qwen3.5-35B-A3B-Base}} and
post-trained\footnote{\url{https://huggingface.co/Qwen/Qwen3.5-35B-A3B}}
Qwen3.5-35B-A3B releases before pruning and after 50\%
expert, FFN or layer pruning. We also evaluate
Qwen3.5-122B-A10B\footnote{\url{https://huggingface.co/Qwen/Qwen3.5-122B-A10B}}
after 75\% expert pruning, alongside its original model on all four tasks.
To assess combinations of pruning axes, two Qwen3.6-35B-A3B models retain 36 layers
and 128 routed experts per layer selected by RAZOR, with intermediate
widths of 408 and 360. Appendix~\ref{app:mixed-design} reports their
results and parameter counts.

The aggregate mathematics analyses use $n=2048$ responses per problem
and pair each pruned model with its original. They cover expert pruning
from 10\% through 75\% on all three MoE backbones, FFN and layer pruning
on Qwen3.6-35B-A3B and GLM-4.7-Flash, and calibrated FFN, calibrated
layer and tail-drop pruning on the dense models.

Table~\ref{tab:model-architectures} summarizes the ten original models
using official reports, model cards and the evaluated configurations.
Layer counts and dimensions refer to the language decoder, excluding
vision encoders and auxiliary multi-token-prediction (MTP) modules.
Dense and MoE designate the FFN architecture independently of attention
layout. Qwen3.5-35B-A3B-Base is pretrained; the other releases are post-trained.

\begin{table}[!htb]
\centering
\small
\setlength{\tabcolsep}{3pt}
\renewcommand{\arraystretch}{1.18}
\begin{tabular*}{\textwidth}{@{\extracolsep{\fill}}llrrrrr@{}}
\toprule
\multicolumn{7}{@{}l}{\textbf{(a) Nominal parameter scales and decoder dimensions}}\\
\midrule
\textbf{Model} & \textbf{FFN type} & \textbf{Total (B)} & \textbf{Active (B)} & \textbf{Layers} & $d_{\rm model}$ & $d_{\rm FFN}$\\
\midrule
Qwen3.6-35B-A3B       & MoE   & 35   & 3   & 40 & 2048 & 512\\
GLM-4.7-Flash         & MoE   & 30   & 3   & 47 & 2048 & 1536\\
Gemma-4-26B-A4B-it    & MoE   & 26   & 4   & 30 & 2816 & 704\\
Qwen3.5-35B-A3B      & MoE   & 35   & 3   & 40 & 2048 & 512\\
Qwen3.5-35B-A3B-Base & MoE   & 35   & 3   & 40 & 2048 & 512\\
Qwen3.5-122B-A10B    & MoE   & 122  & 10  & 48 & 3072 & 1024\\
Qwen3.5-9B           & Dense & 9    & --- & 32 & 4096 & 12288\\
Qwen3.5-4B           & Dense & 4    & --- & 32 & 2560 & 9216\\
Qwen3.5-2B           & Dense & 2    & --- & 24 & 2048 & 6144\\
Llama-3.1-8B-Instruct & Dense & 8    & --- & 32 & 4096 & 14336\\
\bottomrule
\end{tabular*}

\medskip
\begin{tabular*}{\textwidth}{@{\extracolsep{\fill}}lrrrrr@{}}
\toprule
\multicolumn{6}{@{}l}{\textbf{(b) Expert routing and shared FFN paths}}\\
\midrule
& & \multicolumn{2}{c}{\textbf{Routed experts}} & \multicolumn{2}{c}{\textbf{Shared paths}}\\
\cmidrule(lr){3-4}\cmidrule(l){5-6}
\textbf{Model} & \textbf{MoE layers} & \textbf{Experts/layer} & \textbf{top-$k$} & \textbf{Paths/layer} & \textbf{Width}\\
\midrule
Qwen3.6-35B-A3B    & 40 & 256 & 8 & 1 & 512\\
GLM-4.7-Flash      & 46 & 64  & 4 & 1 & 1536\\
Gemma-4-26B-A4B-it & 30 & 128 & 8 & 1 & 2112\\
Qwen3.5-35B-A3B      & 40 & 256 & 8 & 1 & 512\\
Qwen3.5-35B-A3B-Base & 40 & 256 & 8 & 1 & 512\\
Qwen3.5-122B-A10B    & 48 & 256 & 8 & 1 & 1024\\
\bottomrule
\end{tabular*}

\medskip
\begin{minipage}{\textwidth}
A dash means that a separate active-parameter
scale is not reported for the dense model. GLM-4.7-Flash has one initial
dense FFN of width 10240 followed by 46 MoE layers. For all MoE models,
$d_{\rm FFN}$ denotes routed-expert width. Routed top-$k$ excludes the shared
path and is distinct from decoding top-$k$.
\end{minipage}
\caption{\textbf{Original model structure.} Parameters are in billions;
$d_{\rm model}$ and $d_{\rm FFN}$ denote residual and FFN intermediate widths.
Panel (b) lists routing and shared paths.}
\label{tab:model-architectures}
\end{table}

\subsection{Pruning methods and structural ratios}
\label{app:operators}

Expert pruning uses RAZOR~\citep{song2026razor}, while FFN pruning
ranks intermediate channels by first-order Taylor saliency
\citep{ma2023llmpruner}. Layer pruning uses iterative perplexity-based
selection following SLEB~\citep{song2024sleb}, with removal units
matched to the architecture.

The EAN and FREQ baselines follow activation-norm and routing-frequency
pruning~\citep{ffe2025}, using the RAZOR implementations.\footnote{\url{https://github.com/nick7nlp/RAZOR}}
Our EAN variant averages each expert's output $\ell_2$ norm over the
tokens routed to it, while FREQ counts routing selections.
Other controls retain experts by original index order. Tail-drop removes
late layer groups while retaining the final group.

The pruning ratio describes the fraction of structural units removed,
which differs from the fraction of total parameters removed. Unless
labeled actual, percentages refer to the requested ratio $r\in[0,1]$.
For expert pruning, $r_{\rm expert}=1-E_{\rm kept}/E_{\rm original}$
within each MoE layer. Only routed experts enter this count and are
eligible for removal, so shared experts and dense FFNs remain intact.

FFN pruning removes intermediate channels within each dense FFN, routed
expert or shared FFN. Its per-module pruning ratio is
$r_{{\rm FFN},u}=1-d_{u,\rm kept}/d_{u,\rm original}$, while residual width
$d_{\rm model}$ remains unchanged. Layer pruning removes decoder layers,
with $r_{\rm depth}=1-L_{\rm kept}/L_{\rm original}$. Its denominator
includes all original decoder layers, including those ineligible for
removal. Auxiliary prediction heads are counted separately.

Integer rounding can make realized ratios differ slightly from nominal
ones. For example, nominal 10\% expert pruning on Gemma-4-26B-A4B-it
removes 13 of 128 routed experts per layer, or approximately 10.16\%.

To compare compression across pruning axes, we count the remaining model
parameters from tensor shapes and retained module dimensions.
Total parameters include embeddings and the output head,
with tied weights counted once, and exclude vision modules, auxiliary
prediction modules and nonparameter buffers. Active parameters include
attention, normalization, routing, always-active FFNs, selected routed
experts and the full output head. They exclude the input embedding
lookup unless its weights are tied to the output head.

This accounting separates stored parameters from those active per token.
Expert pruning preserves routed top-$k$, so active parameter counts
fall only slightly as the router shrinks. FFN pruning narrows active
modules, while layer pruning removes both attention and FFN parameters.
Consequently, expert and FFN pruning can leave similar total parameter
counts but different active counts, as at 50\% pruning
(Table~\ref{tab:moe-coverage}). The comparison in
Section~\ref{sec:collapses} therefore matches total model size without
matching active computation.

\subsection{Sampling, grading and measurement}
\label{app:sampling}
\label{app:metrics}

We use temperature $0.7$, top-$p$ $0.95$, decoding top-$k$ $20$,
and an 8,192-token cap by default. We estimate pass@$k$ at different
sampling budgets from a common candidate pool for each model and problem
(Eq.~\ref{eq:passk}). Generation settings can differ
between models, as specified below and with the corresponding comparisons.

The mathematics evaluation includes all 30 AIME problems and all
272 Minerva problems, with 2,048 candidates per problem by default.
For Minerva, differences in answer normalization and adjudication affect
which comparisons use a common grading procedure. The paired
Base/Instruct comparison uses a common grader. The mixed-axis
evaluations include additional adjudication, while the Gemma-4-26B-A4B-it
evaluations differ in their treatment of equivalent percentage answers.
Some judgments in the latter evaluations remain ambiguous. Comparisons
across these grading conditions therefore do not isolate the effect of
pruning or expert selection.

The grading procedure also determines how to interpret timeouts and
incomplete generations. In the Minerva tail-drop evaluations, judgments
that exceed the 7,200-second grading limit receive a zero score; this
convention does not establish that the answer is mathematically incorrect.
More generally, the grader evaluates the completion text rather than its
termination status. A response that reaches the length cap may still
contain a gradeable answer, and a normally terminated response may fail
to provide one.

Majority voting selects the most frequent normalized mathematical
answer, excluding unparseable answers and breaking ties randomly.
At smaller budgets, accuracy is estimated from 1,000 resamples drawn
without replacement from the candidate pool.

HumanEval+~\citep{liu2023evalplus} comprises 164 problems evaluated with
the base-plus test suite. The LiveCodeBench-v6 subset
\citep{jain2025livecodebench} contains 175 problems dated January 4 to
April 6, 2025, including 112 standard-input and 63 function-based problems,
with 7,000 tests. The default code budget is $n=K=512$;
other comparison budgets are specified in the corresponding tables and figures.
We omit code majority voting because text agreement does not establish
program equivalence.

Generation settings vary within the original Gemma-4-26B-A4B-it
LiveCodeBench-v6 evaluation. Execution limits also vary, with six seconds per
test and 120 seconds overall in one configuration. Programs that exceed
the execution limit count as failures; evaluation errors are distinguished
from program failures.

We distinguish the available sample count $n$ (or $n_q$ when it varies
by problem), the selected budget $k$ and the comparison endpoint $K$.
Pass@$k$ and majority-vote accuracy are percentages; their differences
are percentage points (pp). The retention measures $E_m$ and $R_{m,K}$
in Eq.~\ref{eq:endpoints} are dimensionless ratios relative to the
same backbone's original model, displayed as ratios in figures or
percentages in tables. A zero denominator makes a ratio undefined;
dashes denote undefined or inapplicable quantities.

Same-$k$ retention compares original and pruned models at equal sampling budgets.
To ask how much sampling a pruned model needs to reach a fixed level of
the original model's coverage, we instead define the target $y_\alpha=\alpha P_0(K)$
and the smallest sampling budget that reaches it:
\begin{equation}
k_m^*(\alpha)=\min\{k\in\{1,\ldots,K\}:P_m(k)\ge y_\alpha\}.
\label{eq:kstar}
\end{equation}
The primary target uses $\alpha=0.9$, with $0.75$ and $1.0$ as
sensitivity checks. Targets not attained within the evaluated range are
right-censored at $K$. If both models attain a target $y$, the empirical
sampling multiplier is $S_m(y)=k_m(y)/k_0(y)$, using the smallest
integer budget for each model. Integer rounding can change this ratio,
so fluctuations at small budgets need not indicate differences in curve
shape.

The presence of a correct candidate does not ensure that an answer
selection rule will recover it. When $P_m(K)>P_m(1)$, we use the gap
between coverage and single-sample accuracy to assess how much of this
available improvement majority voting realizes. The recovered fraction is
\begin{equation}
X_{m,K}=\frac{\majk{K}_m-P_m(1)}{P_m(K)-P_m(1)}.
\label{eq:extract}
\end{equation}
The ratio $X$ can be negative if majority voting lowers accuracy. It is
undefined when coverage equals single-sample accuracy, or for code,
where we do not evaluate majority voting. Interpreting $X$ requires all three scores and the original model's
value because similar fractions can accompany very different
accuracy and coverage.

\section{Reasoning Coverage Across Structures and Models}
\label{app:global}
\label{app:curves}

\subsection{Coverage across pruning structures, backbones and tasks}
\label{app:primary-curves}
\label{app:moe-profiles}
\label{app:four-task-sweeps}
\label{app:backbone-sweeps}
\label{app:code}
\label{app:code-endpoints}
\label{app:decisions}

\begin{figure}[!t]
\centering
\includegraphics[width=\textwidth]{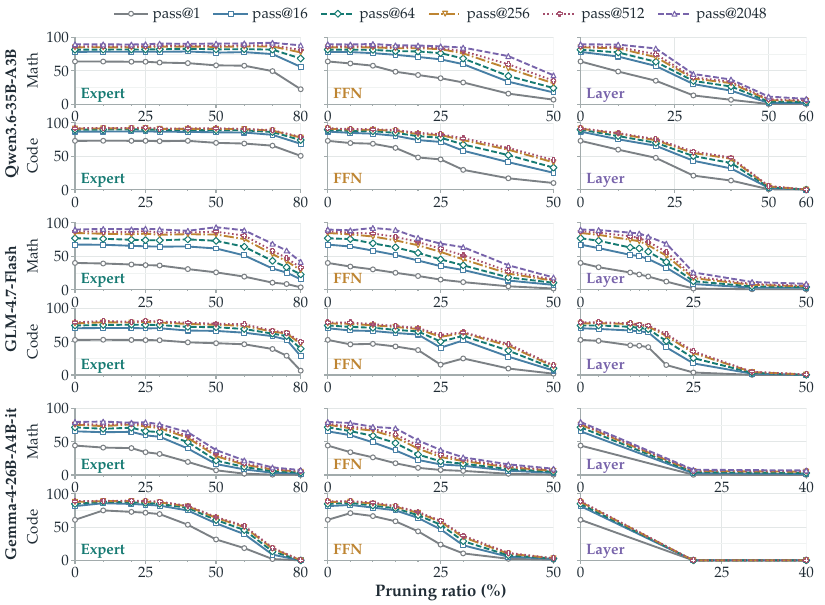}
\caption{\textbf{Math and code coverage across pruning ratios.} Columns compare expert, FFN and layer pruning; 0\% denotes the original model. For each backbone, Math averages AIME-2026 and Minerva ($k\leq2048$); Code averages HumanEval+ and LiveCodeBench-v6 ($k\leq512$). Curves show mean pass@$k$ with equal task weights.}
\label{fig:severity}
\end{figure}

Figure~\ref{fig:severity} shows how the benefit of additional sampling
depends on what is pruned and how much is removed. We plot mathematics
and code separately to expose differences hidden by a four-task average.

\begin{figure}[!t]
\centering
\includegraphics[width=\textwidth]{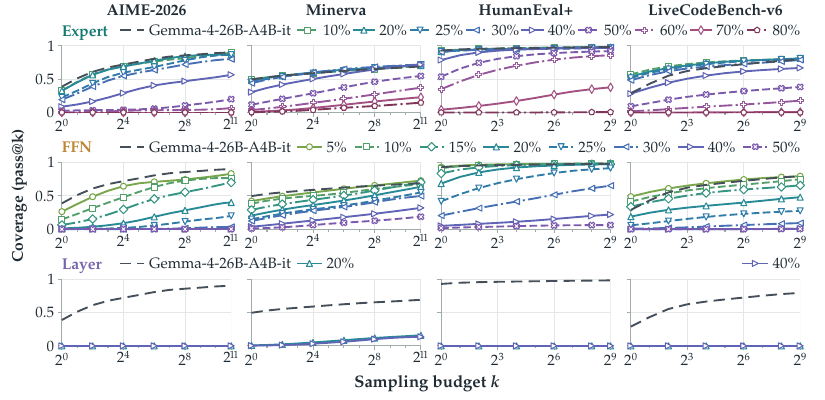}
\caption{\textbf{Gemma-4-26B-A4B-it coverage across four tasks.} Rows show expert, FFN and layer pruning. We use $n=2048$ for mathematics and $n=512$ for code. Sampling and grading conditions are given in Appendix~\ref{app:sampling}.}
\label{fig:fourtask-gemma}
\end{figure}

Increasing the sampling budget does not compensate for every pruning
structure or severity. Gemma-4-26B-A4B-it makes the distinction clear.
Mild expert pruning retains much of its coverage
despite lower accuracy, whereas severe pruning reduces both. Layer
pruning reaches this failure regime much earlier: at both 20\% and 40\%,
the model produces no successful AIME or code response within the
evaluated budgets (Figure~\ref{fig:fourtask-gemma}). We compare pruning
structures within the same backbone, since expert pruning applies only
to MoEs.

No single expert-pruning ratio separates coverage retention from
collapse across backbones. At 80\%, Qwen3.6-35B-A3B still covers 27
AIME problems against the original model's 29, whereas Gemma-4-26B-A4B-it has
no observed success with 2,048 samples per problem. Generation
conditions differ, and the original Gemma-4-26B-A4B-it has a high
truncation rate, so this contrast cannot be attributed solely to
architecture. Retaining at least the routed top-$k$ experts ensures
feasible routing; it does not by itself establish a safe pruning ratio.

Even within a backbone, the apparent limit depends on the task.
Gemma-4-26B-A4B-it retains nonzero Minerva coverage after layer
pruning has eliminated observed success on AIME and code.
Under severe expert pruning, GLM-4.7-Flash preserves HumanEval+
coverage substantially better than AIME or LiveCodeBench-v6.
The difference between tasks persists at a common sampling budget:
at $k=64$, Gemma-4-26B-A4B-it expert-50\% retains 91.6\% of the
original HumanEval+ score but only 7.3\% of its AIME score.
Different maximum budgets do not fully explain this contrast. Assessing
retained coverage requires specifying both the task and the budget;
an average can conceal severe losses on an individual benchmark.

\begin{figure}[!t]
\centering
\includegraphics[width=\textwidth]{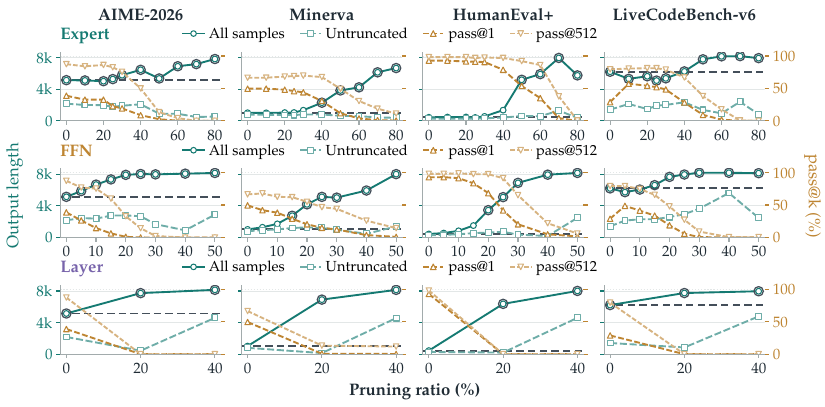}
\caption{\textbf{Output length, accuracy and coverage on Gemma-4-26B-A4B-it.}
Rows show expert, FFN and layer pruning across four tasks; 0\% denotes
the original model. Left axes show mean tokens for all and untruncated
samples; right axes show pass@1 and pass@512. Grey dashed lines mark
original output length; rings mark $\geq20\%$ truncation.}
\label{fig:gemma-length}
\end{figure}

Changes in output length do not consistently track coverage retention.
At 30\% expert pruning, Qwen3.6-35B-A3B covers 146
LiveCodeBench-v6 problems against the original model's 148, with
essentially unchanged length. At 80\%, output becomes shorter while
accuracy and coverage deteriorate; coverage falls from 84.6\% to
60.6\% at $K=512$. On Gemma-4-26B-A4B-it, 30\% FFN pruning instead
lengthens Minerva outputs from 1,021 to 5,046 tokens while pass@1 falls
from 49.6\% to 12.3\% (Figure~\ref{fig:gemma-length}). Longer output
can thus accompany declining success, while shorter output need not
indicate recovery. To understand which problems are more vulnerable,
we next examine their original response lengths while accounting for
differences in original success rates.

\subsection{Task demands and pruning vulnerability}
\label{app:task-length}
\label{app:within-task}
\label{app:within-task-cost}

The cross-task curves suggest greater pruning vulnerability on tasks where
the original model generates longer responses, but length and original accuracy
need not rank tasks in the same way. Qwen3.6-35B-A3B, for example,
generates longer responses on AIME than on Minerva while answering
AIME more accurately. Across tasks, longer original responses generally
accompany lower pass@512 retention under FFN and layer pruning, although
expert pruning of Qwen3.6-35B-A3B is a prominent exception
(Figure~\ref{fig:length}). This contrast motivates a within-task comparison to examine whether
longer original responses are associated with greater pruning vulnerability
among problems with similar original success rates.

We compare problems within each task to reduce differences in answer
format across tasks. Original accuracy controls one
observable aspect of difficulty, while output length captures the amount
of generation used to produce an answer. Both are imperfect difficulty
measures, since accuracy uses a finite pool and length also reflects
verbosity and truncation.

We select problems answered correctly in every original sample, using
512 samples per problem for code and 2,048 for mathematics. Within each
backbone and task, we rank them by the original model's mean response length and
split them into equal shorter and longer halves, omitting the middle
problem when the count is odd. Each half has at least five problems
and 100\% empirical original pass@1, and remains fixed across pruning ratios.

At 30\% FFN pruning, the longer half has lower pass@1 in each comparison
shown in Table~\ref{tab:within-task-all-correct}. This direction
holds from 5\% to 50\% FFN pruning for the eligible code groups
and Qwen3.6-35B-A3B Minerva, with small reversals for
GLM-4.7-Flash Minerva at mild pruning. GLM-4.7-Flash LiveCodeBench-v6
and both AIME comparisons lack enough all-correct problems to qualify.

\begin{table}[!htbp]
\centering
\renewcommand{\arraystretch}{1.18}
\small
\setlength{\tabcolsep}{3.5pt}
\begin{tabular*}{\textwidth}{@{\extracolsep{\fill}}llrrr@{}}
\toprule
\textbf{Model} & \textbf{Task} & $m$ & \textbf{Original tokens} & \textbf{Pruned pass@1 (\%)} \\
 & & & \textbf{short / long} & \textbf{short / long} \\
\midrule
\multicolumn{5}{@{}l}{\textbf{FFN pruning: 30\%}}\\
\addlinespace[2pt]
Qwen3.6-35B-A3B & HumanEval+ & 60 & 137 / 286 & 66.19 / 47.88 \\
GLM-4.7-Flash & HumanEval+ & 20 & 105 / 217 & 88.66 / 61.27 \\
Qwen3.6-35B-A3B & Minerva & 35 & 464 / 898 & 95.87 / 82.68 \\
GLM-4.7-Flash & Minerva & 8 & 283 / 496 & 82.46 / 58.59 \\
\bottomrule
\end{tabular*}
\caption{\textbf{Different pruning losses despite 100\% observed original accuracy.}
Each half contains $m$ problems; tokens are mean full-output lengths.}
\label{tab:within-task-all-correct}
\end{table}

Restricting the analysis to problems with perfect observed original
accuracy leaves some comparisons with too few eligible problems.
To examine whether length also distinguishes pruning vulnerability
beyond this subset, we group problems with at least one original
success into frequency bands of two, five and ten percentage points.
Within each band, we split problems by original mean output length,
omit an unpaired middle problem, and pool the shorter and longer
halves across bands. Small differences in original accuracy remain
within a band, so we compare the two groups' losses in pass@1 relative
to their own original accuracy.

On both code tasks, the longer group loses more accuracy in nearly all
tested combinations of FFN pruning ratio and frequency-band width
(Table~\ref{tab:within-task-stratified}). The mathematics results are
less consistent, especially on AIME-2026, so the association between
original response length and subsequent accuracy loss varies by task
even after grouping problems by empirical original accuracy. This
grouping controls only part of the variation in difficulty and does
not establish a causal effect of length. All pruning ratios reuse the
same problems, so the pattern across ratios describes how those
problems respond to different pruning severities.

\begin{table}[!htbp]
\centering
\renewcommand{\arraystretch}{1.18}
\small
\begin{tabular*}{\textwidth}{@{\extracolsep{\fill}}lrrr@{}}
\toprule
\textbf{Task} & \textbf{2-point bands} & \textbf{5-point bands} & \textbf{10-point bands} \\
\midrule
HumanEval+ & 16/16 (15.13) & 16/16 (15.36) & 16/16 (16.01) \\
LiveCodeBench-v6 & 16/16 (8.08) & 16/16 (9.49) & 15/16 (10.98) \\
Minerva & 11/16 (2.92) & 16/16 (4.79) & 15/16 (4.08) \\
AIME-2026 & 8/16 (0.46) & 14/16 (3.90) & 14/16 (5.67) \\
\bottomrule
\end{tabular*}
\caption{\textbf{Length-related FFN pruning losses at similar original success frequencies.}
Cells count configurations in which the longer half loses more accuracy,
followed by the median extra loss in percentage points. Each task covers
two backbones and eight ratios.}
\label{tab:within-task-stratified}
\end{table}

To connect these accuracy losses to generation cost, we use tokens per
correct sample, dividing all generated tokens, including failed attempts,
by the number of correct samples. This measures aggregate token expenditure rather than the
expected cost of solving each problem or the cost of reaching a fixed
coverage target (Appendix~\ref{app:cost}).

The length-related cost increase in Section~\ref{sec:task-demands}
occurs even among problems that the original model solves reliably.
For GLM-4.7-Flash on LiveCodeBench-v6, we select problems with at least
90\% original success frequency, split them equally by original mean
output length and keep the groups fixed across pruning ratios, using
512 samples per problem. At 30\% FFN pruning, all 14 problems in the
longer group remain covered despite the sharp accuracy loss. Truncation
reaches 19.14\%, raising the question of whether length limits alone
explain that loss.

To assess whether truncation accounts for these failures, we bound the
fraction of all outputs that are incorrect without being length-truncated.
For correct fraction $p_{\mathrm{correct}}$ and length-truncated fraction
$p_{\mathrm{trunc}}$, this fraction is at least
$\max\{0,1-p_{\mathrm{correct}}-p_{\mathrm{trunc}}\}$.
This bound allows correct and truncated samples to overlap.
Among LiveCodeBench-v6 problems that Qwen3.6-35B-A3B answers correctly
in every original sample, the longer-response half has 29.94\% pass@1
and 2.08\% truncation after 30\% FFN pruning. At least 67.98\% of all
outputs are therefore incorrect and not length-truncated. The
GLM-4.7-Flash group above gives a bound of 50.95\%.
Thus, failures on these originally reliable problems cannot all be
explained by a hard length cap. The bound does not distinguish reasoning
errors, incoherent outputs or other forms of failure.

\subsection{FFN and layer pruning in dense models}
\label{app:dense-main-layout}

\begin{figure}[!t]
\centering
\includegraphics[width=\textwidth]{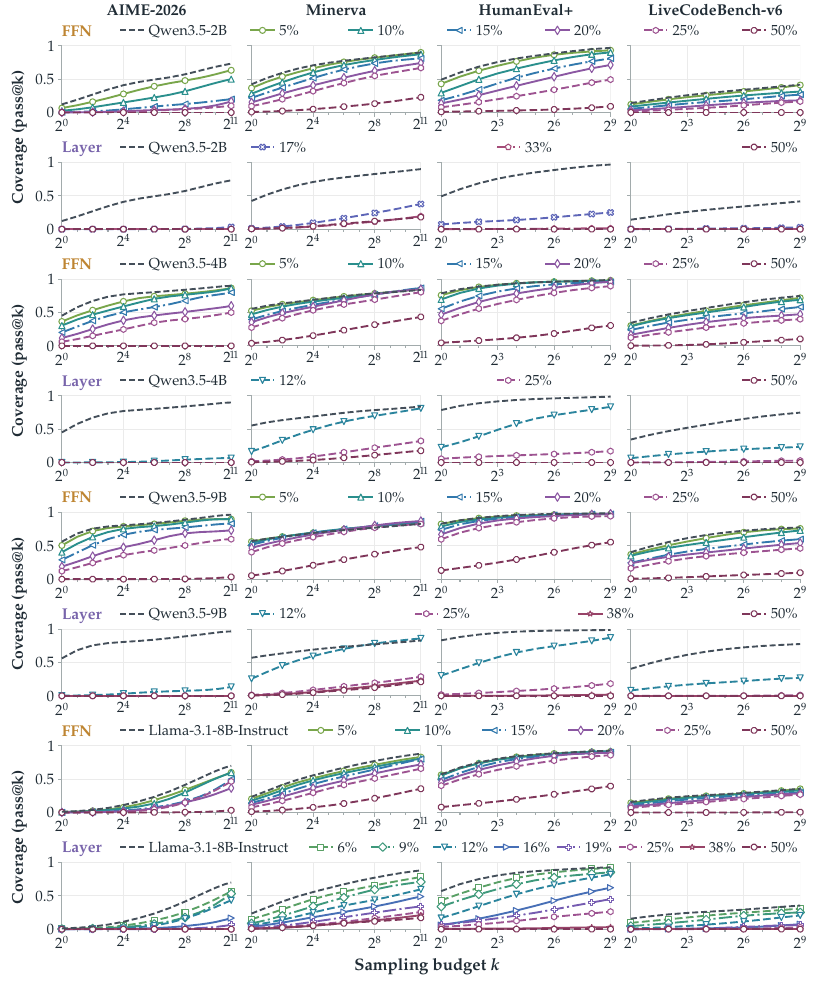}
\caption{\textbf{Coverage after FFN and layer pruning in dense models.} Paired rows show Qwen3.5-2B, Qwen3.5-4B, Qwen3.5-9B and Llama-3.1-8B-Instruct, each compared with its original model. Mathematics reaches $k=2048$ and code $k=512$. Zero curves indicate no successful response in the generated pool.}
\label{fig:dense-passk}
\end{figure}

In dense models, the coverage recovered through additional sampling
also depends on which structures are retained
(Figure~\ref{fig:dense-passk}). Pruning 25\% of Qwen3.5-4B's FFN width
retains more coverage than pruning 25\% of its layers on both mathematics
tasks, although both remain below the original model. At 50\% pruning, both axes have
zero observed AIME coverage with 2,048 samples per problem.
These Qwen3.5-4B comparisons test robustness within the model family:
removing 50\% of its FFN width does not reproduce the original
Qwen3.5-2B width.

\begin{figure}[!t]
\centering
\includegraphics[width=\textwidth]{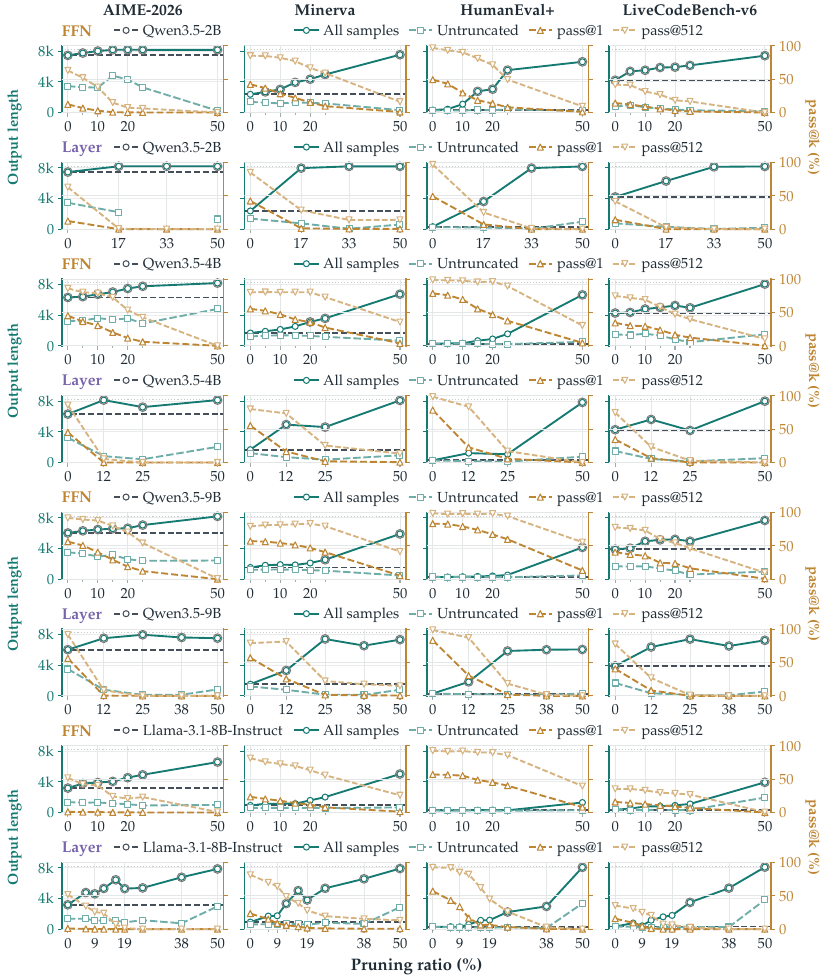}
\caption{\textbf{Output length and success after pruning in dense models.} Paired rows show Qwen3.5-2B, Qwen3.5-4B, Qwen3.5-9B and Llama-3.1-8B-Instruct. Left axes show mean lengths over all responses and over untruncated responses; right axes show pass@1 and pass@512. The grey dotted line marks the 8,192-token cap; grey rings mark truncation rates of at least 20\%.}
\label{fig:dense-length}
\end{figure}

The length measurements show that the loss of observed successes
can coexist with longer outputs (Figure~\ref{fig:dense-length}).
For Qwen3.5-4B on AIME at 50\% FFN or layer pruning, the mean over all
responses approaches the token cap, while the mean over normally
terminated responses remains much lower. Shorter responses in this
subset do not imply lower overall token use.

\subsection{Selection criteria and sampling efficiency}
\label{app:selection}
\label{app:selection-controls}

Selection can affect how much coverage pruning preserves. At a nominal
layer pruning ratio of 4\%, GLM-4.7-Flash covers 10.0\% of AIME under tail-drop and
90.0\% under calibrated selection at $K=2048$. At 50\% expert
pruning, Qwen3.6-35B-A3B's positional control covers 43.3\%, compared
with 96.7\% for each of RAZOR, EAN and FREQ at the same budget.

The positional controls thus retain substantially less coverage than
calibrated selection. Among RAZOR, EAN and FREQ,
the AIME comparison instead concerns how efficiently the same solved
set is recovered (Section~\ref{sec:selection-efficiency}). On both code
tasks, pass@1 differences also exceed coverage differences at the
common $K=512$ budget, although the endpoint coverage is not identical.

Attempt counts alone do not determine generated-token use because
the criteria also differ in mean response length, especially at
50\% pruning on Minerva (Figure~\ref{fig:expert-strategy}). The
mean over all responses captures the cost per attempt, including failures and
capped responses; the untruncated mean describes only responses that
terminate normally. The fixed-target token measure
(Section~\ref{sec:selection-work}) multiplies the required attempts by
the mean length over all responses.

\subsection{Training stage and model size}
\label{app:inheritance}
\label{app:figure-length-bands}

Higher coverage before pruning does not guarantee higher coverage
afterward. In the Qwen3.5-35B-A3B Base/Instruct comparison, Base leads
on Minerva before pruning and after 50\% expert pruning, but Instruct
leads after 50\% FFN pruning (Table~\ref{tab:training-stage}).
The relative performance of the two releases therefore depends on
the retained structure. These comparisons match pruning axes and
nominal ratios and use a common Minerva grader; they do not isolate
individual post-training stages.

Relative to their own original models, both releases preserve or
exceed AIME and Minerva coverage under expert pruning, whereas FFN
and layer pruning reduce coverage. This retention also depends on
the task: Instruct expert-50\% gains one covered AIME problem despite
lower pass@1, but loses coverage on LiveCodeBench-v6.

\begin{table}[!htbp]
\centering
\renewcommand{\arraystretch}{1.18}
\small
\setlength{\tabcolsep}{3pt}
\begin{tabular*}{\textwidth}{@{\extracolsep{\fill}}llcccc@{}}
\toprule
\textbf{Release} & \textbf{Setting} & \textbf{AIME-2026} & \textbf{Minerva} & \textbf{HumanEval+} & \textbf{LiveCodeBench-v6}\\
\midrule
Base & Original & $35.33\,/\, 80.00$ & $49.07\,/\, 85.66$ & $82.77\,/\, 98.78$ & $33.12\,/\, 76.57$\\
Base & Expert 50\% & $35.60\,/\, 80.00$ & $47.99\,/\, 88.97$ & $82.38\,/\, 98.78$ & $31.09\,/\, 76.57$\\
Base & FFN 50\% & $2.71\,/\, 40.00$ & $13.79\,/\, 58.09$ & $14.81\,/\, 70.12$ & $2.72\,/\, 20.57$\\
Base & Layer 50\% & $<0.01\,/\, 6.67$ & $0.45\,/\, 19.85$ & $0.53\,/\, 10.37$ & $0.00\,/\, 0.00$\\
\midrule
Instruct & Original & $61.36\,/\, 93.33$ & $59.29\,/\, 79.04$ & $88.95\,/\, 98.17$ & $51.22\,/\, 84.00$\\
Instruct & Expert 50\% & $48.99\,/\, 96.67$ & $55.74\,/\, 81.62$ & $90.47\,/\, 98.17$ & $52.51\,/\, 81.71$\\
Instruct & FFN 50\% & $6.10\,/\, 53.33$ & $25.60\,/\, 70.22$ & $27.38\,/\, 81.10$ & $6.59\,/\, 26.86$\\
Instruct & Layer 50\% & $<0.01\,/\, 6.67$ & $0.56\,/\, 21.32$ & $0.84\,/\, 7.93$ & $0.00\,/\, 0.00$\\
\bottomrule
\end{tabular*}
\caption{\textbf{Pruning at two training stages of Qwen3.5-35B-A3B.}
Cells give pass@1 / pass@$K$ (\%); $K=2048$ for math and $512$
for code. Positive scores below 0.005\% are shown as $<0.01$.}
\label{tab:training-stage}
\end{table}

An advantage over a smaller original model is informative only in
relation to the resources retained. With 5\% FFN pruning,
Qwen3.5-9B exceeds Qwen3.5-2B in coverage on AIME and both code
tasks, but trails on Minerva (Section~\ref{sec:native-small}).
The lightly pruned model still retains most of the larger model's FFN
width, so its coverage advantage alone does not establish greater
resource efficiency.

Matching FFN width exposes the limit of that advantage
(Table~\ref{tab:matched-width}). Pruning 25\% or 50\% of
Qwen3.5-9B's FFN width matches Qwen3.5-4B or Qwen3.5-2B,
respectively, but yields lower accuracy and coverage on every task.
The AIME coverage gaps are substantial: 60.0\% versus 90.0\%
at the Qwen3.5-4B width, and 3.3\% versus 73.3\% at the
Qwen3.5-2B width. Retaining a larger model's representations does
not compensate for the width reduction in these comparisons.
Other architectural dimensions, total parameters and training compute
remain unmatched.

\begin{table}[!htbp]
\centering
\renewcommand{\arraystretch}{1.18}
\small
\setlength{\tabcolsep}{3pt}
\begin{tabular*}{\textwidth}{@{\extracolsep{\fill}}lcccc@{}}
\toprule
\textbf{Model} & \textbf{AIME-2026} & \textbf{Minerva} & \textbf{HumanEval+} & \textbf{LiveCodeBench-v6}\\
\midrule
Qwen3.5-9B, FFN 25\% & $12.07\,/\, 60.00$ & $40.51\,/\, 82.72$ & $59.83\,/\, 94.51$ & $16.30\,/\, 46.29$\\
Original Qwen3.5-4B & $45.16\,/\, 90.00$ & $55.45\,/\, 83.82$ & $78.71\,/\, 98.78$ & $34.50\,/\, 74.86$\\
\midrule
Qwen3.5-9B, FFN 50\% & $<0.01\,/\, 3.33$ & $5.36\,/\, 48.16$ & $13.08\,/\, 55.49$ & $0.65\,/\, 9.71$\\
Original Qwen3.5-2B & $12.22\,/\, 73.33$ & $42.31\,/\, 90.07$ & $49.23\,/\, 96.95$ & $14.19\,/\, 41.71$\\
\bottomrule
\end{tabular*}
\caption{\textbf{Pruned Qwen3.5-9B and smaller original Qwen3.5 models at matched FFN widths.}
Cells give pass@1 / pass@$K$ (\%), with $K=2048$ for mathematics and
$K=512$ for code. Total parameters and training compute are not matched.}
\label{tab:matched-width}
\end{table}

The Qwen3.5-122B-A10B comparison extends this analysis to expert
pruning at a larger model scale (Figure~\ref{fig:122b}).
At 75\% expert pruning, the model preserves its
original AIME coverage and slightly increases its Minerva coverage,
even though pass@1 decreases. Coverage falls on both code tasks,
with the larger loss on LiveCodeBench-v6 at $K=256$.

Against the smaller original Qwen3.5-35B-A3B, the same pruned model
matches AIME coverage and slightly exceeds Minerva
coverage, but covers fewer problems on both code tasks. The comparison
with Qwen3.5-35B-A3B-Base changes which tasks favor the pruned model:
it covers more AIME problems but fewer problems on each of the other
three tasks. Thus, the preservation of the larger source model's
mathematics coverage does not translate into a consistent advantage
across tasks and smaller original models, whose training histories and
compute budgets also differ.

The models in Figure~\ref{fig:122b} use the sampling protocol in
Appendix~\ref{app:sampling} with repetition penalty 1.0. The prompt
template requests generation without thinking, but model releases may
respond differently to this request.

The length curves also show why shorter completed responses need not
indicate more efficient reasoning. On LiveCodeBench-v6, pruning
Qwen3.5-122B-A10B increases the mean over all outputs but decreases
the mean among untruncated outputs as truncation becomes more frequent.
The latter describes a changing subset: long attempts that hit the cap
are excluded. The pale segment in Figure~\ref{fig:122b} marks the gap
between these two means, making this selection effect visible.

\section{Generation Efficiency, Reliability and Problem-Level Changes}
\label{app:local}
\label{app:items}

\subsection{Sampling, answer selection and token cost}
\label{app:availability}
\label{app:fixed-targets}
\label{app:majority-extraction}
\label{app:cost}

The practical value of retained coverage depends on the cost of producing
correct candidates and the ability to select them. We examine these
requirements in turn, starting with the generation needed to reach a
common coverage target.

At budget $k$, estimated generation is $k\bar t_m$ tokens per problem,
where $\bar t_m$ includes failed and truncated responses. The target
budgets in Eq.~\ref{eq:kstar} assign the same number of candidates to
each problem, without stopping after success. This measure separates
costly recovery of coverage from failure to reach the target.
Unattained targets remain censored at the
sampling cap.

On AIME-2026, paired bootstrap resampling of problems assesses the
uncertainty in these target-based comparisons. With 75\% expert pruning,
Qwen3.6-35B-A3B requires an estimated 246 samples for the same relative
target, with a 95\% interval of 23--824. All 4,000 resamples reach
the target, so the uncertainty here concerns the required sampling
budget. After 40\% FFN pruning, Qwen3.6-35B-A3B misses the target in
nearly all resamples, and after 50\% expert pruning,
Gemma-4-26B-A4B-it misses it in every resample.

\begin{figure}[t]
\centering
\includegraphics[width=\textwidth]{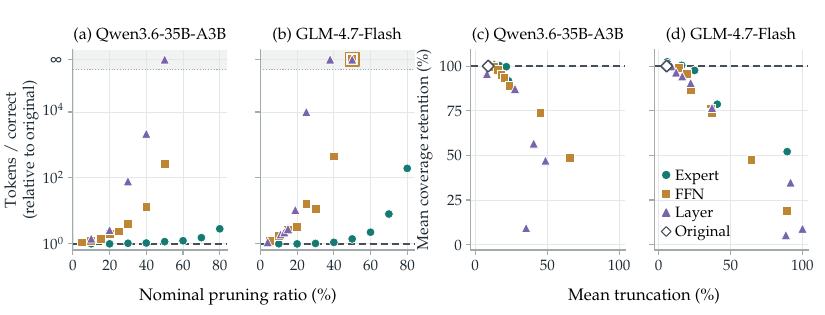}
\caption{\textbf{Token requirements and truncation under pruning.}
(a,c) Qwen3.6-35B-A3B; (b,d) GLM-4.7-Flash.
All panels average equally over AIME-2026, Minerva, HumanEval+ and
LiveCodeBench-v6. (a,b) Tokens per grader-correct completion, normalized
within each task by the original model before averaging; $\infty$ marks
no observed correct completion on at least one task.
(c,d) Task-averaged truncation rates versus coverage retention; only
coverage is normalized by the original model within each task before
averaging. Coverage uses $K=2048$ for mathematics and $K=512$ for code.}
\label{fig:cost}
\end{figure}

Fixed-target token use measures the cost of reaching a specified coverage.
Figure~\ref{fig:cost}(a,b) instead measures the generated tokens used per
grader-correct completion. Within each task, it
combines mean generated length $\bar t_m$ with single-sample accuracy
$\mu_m=P_m(1)$ in fractional units.
No finite observed cost is available when a model has no successful
completion. For $\mu_m>0$, the aggregate token requirement is
\begin{equation}
T_m=\frac{\bar t_m}{\mu_m},\qquad
\frac{T_m}{T_0}=\frac{\bar t_m}{\bar t_0}\frac{\mu_0}{\mu_m}.
\label{eq:cost}
\end{equation}
We average these normalized ratios equally across the four tasks.
This factorization is exact for aggregate tokens per correct completion
within a task.
It need not equal $\Exp_q[t_q/p_q]$, the cost of solving each problem
independently by repeated sampling, which can be dominated by problems
with small $p_q$.

Token comparisons are within each backbone because tokenizers differ.
They exclude prompt processing and verification. Per-token computation
also changes differently across pruning axes (Appendix~\ref{app:operators}),
so token ratios alone do not measure runtime savings.

Even affordable correct candidates are useful only if they can be
identified. Qwen3.6-35B-A3B with 75\% expert pruning illustrates the gap
that remains under majority voting. It retains 96.7\% AIME oracle coverage, yet majority
accuracy is only 66.7\%, compared with 90.0\% for the original model.
The sampled candidates therefore contain correct answers more often
than majority voting selects one. The extraction ratio in
Eq.~\ref{eq:extract} measures how much of the gap between single-sample
accuracy and coverage is recovered by majority voting. It should be read
alongside those scores, especially when the gap is small.

\subsection{Generation reliability and diversity of generated CoTs}
\label{app:behavior}
\label{app:truncation}
\label{sec:termination-reliability}
\label{sec:strategy-diversity}

To understand why longer outputs need not improve coverage, we examine
whether responses terminate, how much they repeat and how they differ
across samples. Truncation is the fraction of samples stopped by the token
cap. It captures one failure mode, but a normally terminated output can
be incorrect and a capped output may already contain a correct answer.
Qwen3.5-9B with 50\% tail-drop, for example, has only 7.0\% AIME truncation
and no observed success among $n=2048$ candidates per problem. Low
truncation can thus accompany complete failure.

Figure~\ref{fig:process} summarizes repetition and CoT diversity across
all four tasks, using 2,048 candidates per mathematics problem and 512
per coding problem. The task-level analyses below use AIME and Minerva
with 2,048 candidates per problem. For Qwen3.6-35B-A3B and GLM-4.7-Flash,
RAZOR expert pruning ranges from 10\% to 80\% in increments of
10 percentage points.

The layer-pruning results show how the composition of failures changes.
In the AIME comparisons, falling
truncation accompanies an increase in normally terminated incorrect
responses (Table~\ref{tab:termination-outcomes}). At 50\% layer pruning,
nearly all normally terminated Qwen3.6-35B-A3B responses and all such
GLM-4.7-Flash responses are incorrect. These outcome frequencies do not
establish whether forcing earlier termination would improve correctness.
Appendix~\ref{sec:decoding} examines how changing the token cap and
repetition penalty affects generation.

\begin{table}[!htb]
\centering
\renewcommand{\arraystretch}{1.18}
\small
\setlength{\tabcolsep}{4pt}
\begin{tabular*}{\textwidth}{@{\extracolsep{\fill}}llrrrr@{}}
\toprule
& & \multicolumn{2}{c}{\textbf{Normal termination}} & \multicolumn{2}{c}{\textbf{Token cap}}\\
\cmidrule(lr){3-4}\cmidrule(l){5-6}
\textbf{Backbone} & \textbf{Setting} & \textbf{Correct} & \textbf{Incorrect} & \textbf{Correct} & \textbf{Incorrect}\\
\midrule
Qwen3.6-35B-A3B & Original & 42,778 & 6,322 & 9 & 12,331\\
& Layer 30\% & 223 & 4,153 & 3 & 57,061\\
& Layer 40\% & 7 & 10,098 & 0 & 51,335\\
& Layer 50\% & 2 & 27,939 & 0 & 33,499\\
\midrule
GLM-4.7-Flash & Original & 19,983 & 38,794 & 4 & 2,659\\
& Layer 25\% & 0 & 578 & 1 & 60,861\\
& Layer 38\% & 0 & 4 & 0 & 61,436\\
& Layer 50\% & 0 & 3,287 & 1 & 58,152\\
\bottomrule
\end{tabular*}
\caption{\textbf{Answer correctness and termination under layer pruning.}
Each row contains 61,440 responses: 2,048 candidates for each of 30 AIME
problems. Correctness is determined by the answer grader; normal
termination is determined by the generation finish reason.}
\label{tab:termination-outcomes}
\end{table}

Termination alone also leaves open whether an output uses its length
productively. We therefore distinguish repetition within a response
from diversity across generated CoTs
(Section~\ref{sec:trace-analysis}). Length and repetition summaries
use normally terminated outputs, called complete traces, with
whitespace tokenization. The 4-gram repetition score is one minus the
ratio of distinct to total 4-grams and is undefined for traces shorter
than four tokens. Length and 4-gram repetition are averaged within
each eligible problem and then equally across problems. A separate
indicator records whether an 8-gram occurs at least three times in a
response. This identifies repeated text without establishing a
sustained generation loop. To summarize how frequently this occurs,
Figure~\ref{fig:process} pools all complete responses within each task,
including incorrect responses, before averaging the repeated-8-gram
rates equally across tasks. The separate AIME comparison below
conditions on correct complete traces to examine whether repetition
also occurs among responses that receive a correct answer grade.

Cross-candidate diversity is measured by symmetric clipped 4-gram
overlap: the matched count, clipped by the two occurrence counts,
divided by the smaller 4-gram total. This differs from standard Self-BLEU.
For each problem, eligible outputs are normally terminated responses
with at least four whitespace-delimited tokens among the 2,048 mathematics
or 512 coding candidates, including incorrect responses. We sample up to 1,024
unordered pairs uniformly without replacement, using all pairs when
there are at most 1,024. Pair scores are averaged within each problem,
then equally across eligible problems; fewer than two eligible outputs
makes overlap undefined. Figure~\ref{fig:process} averages the four
task-level scores equally and leaves overlap undefined if any task
has no eligible pair.
Overlap measures textual variation across generated CoTs. Different
wording can express the same strategy, so lower overlap alone does not
establish more diverse valid reasoning.

Responses can become less similar to one another while containing
more repetition individually. In the four-task averages, both backbones
have layer-pruned configurations with lower overlap but more frequent
repetition than their original models (Figure~\ref{fig:process}). The opposite repetition trend is also possible.
For GLM-4.7-Flash at 50\% expert pruning, repetition falls to 16.9\%
and overlap to 0.249, both below their original levels. Lower overlap therefore does not
determine the direction of the repetition change. At high truncation,
these statistics describe only the few outputs that terminate normally,
whose low repetition rate need not indicate reliable generation overall.

Answer correctness does not rule out repeated text within a response
either. On AIME, 88.8\% of correct complete traces from
GLM-4.7-Flash at 19\% layer pruning contain an 8-gram at least
three times, compared with 64.8\% for the original model.

\subsection{Changes in solved problems and success frequencies}
\label{app:solved-sets}
\label{app:frequency-transitions}

Net coverage summarizes how many problems are solved, but not how
pruning changes that set. A gain can coexist with losses on problems
the original model solved. Matching problem identities separates
retained, lost and newly observed successes
(Appendix~\ref{app:curve-information}); a lost success means that the
original candidate pool contains a correct response and the pruned
pool contains none.

At 50\% expert pruning, Qwen3.6-35B-A3B retains the original
AIME solved set. On Minerva, however, it retains 218 problems,
gains 13 and loses eight within 2,048 candidates per problem.
The net gain thus combines new successes with losses from the original solved set.

Solved-set membership records whether success was observed;
per-problem frequencies show how often it occurred. We compare
per-problem success frequencies on the same 30 AIME problems for Qwen3.6-35B-A3B
and three pruned models, each with 2,048 candidates per problem. The four
bands are no observed success, positive frequency below 5\%, 5\% to
below 50\%, and at least 50\%. Movement between positive bands measures
changes in reliability. Entry into the first band records the absence of
an observed success, not a zero underlying success probability.
Of the 22 AIME problems the original model solves in at least half its
candidates, expert-50\% keeps 21 in that frequency band, whereas FFN-40\%
keeps only one. FFN pruning therefore sharply reduces the number of
problems with frequent observed success.

\subsection{Answer correctness and derivation validity}
\label{app:measurement}
\label{app:measurement-scope}
\label{app:reasoning-counterexample}

For one AIME problem, Qwen3.6-35B-A3B with 25\% expert pruning produces
only one accepted answer among 2,048 candidates. This response terminates
normally and boxes 393, but its derivation is invalid. It illustrates why
rare successes require closer examination when interpreting coverage.

The question asks for the number of ordered tuples in $\{1,2,3\}^7$
satisfying $\sum_i a_i\equiv0$ and
$\sum_{i=1}^7 a_i a_{i+1}a_{i+3}\equiv0\pmod3$, with cyclic indices.
Exact integer enumeration of all $3^7$ tuples gives 393 solutions.
Among the 729 tuples satisfying the first constraint, the cubic residues
occur $(393,168,168)$ times.

The candidate uses a roots-of-unity filter to count these tuples.
Writing $x_i=a_i\bmod3$, it evaluates
$Z(u,1)=\sum_{\boldsymbol x}\omega^{u\sum_i x_i+S(\boldsymbol x)}$,
where $\omega$ is a nontrivial cube root of unity and $S$ is the cubic
constraint. The error occurs when summing out $x_1$: the remaining
phase must still include $u\sum_{i=2}^7x_i$, but the candidate drops
this term for $u=1,2$ and repeats the error in its final verification.
Exact evaluation gives $(567,-189,297)$ for $u=(0,1,2)$; the candidate's
expressions instead give $(567,54,54)$. The two errors, $+243$ and
$-243$, cancel in the total, producing the correct answer 393 through
an invalid derivation.

The final-answer grader correctly accepts this response under its
criterion. At $k=2048$, however, this single accepted response contributes
3.33 percentage points to coverage on the 30-problem AIME benchmark.
The example shows why endpoint coverage cannot by itself establish
valid reasoning on every solved problem. It does not estimate the
prevalence of invalid derivations or show that pruning caused this error.

\section{Statistical Support and Robustness}
\label{app:statistics}

\subsection{Coverage changes beyond uniform success loss}
\label{app:thinning}
\label{sec:matched}
\label{app:reference-support}
\label{sec:thinning}
\label{app:thinning-transport}
\label{app:reference-scales}
\label{app:paired-uncertainty}
\label{app:foundations}
\label{app:allocation-theory}

A fall in pass@1 can arise because correct solutions become rarer on
most problems or because success disappears on particular problems.
These changes can have different consequences for coverage even when
the mean loss is the same. We test the first explanation with a
uniform-loss reference: independently retain each original-model success
with probability $a_m=P_m(1)/P_0(1)$. This matches the pruned model's
pass@1 in expectation without fitting its higher-budget coverage.

The two pruning structures depart from this reference differently.
On Qwen3.6-35B-A3B AIME, 75\% expert pruning
is close to the reference at $k=2048$ (96.7\% versus 95.2\%), whereas
40\% FFN pruning falls far below it (63.3\% versus 92.1\%). The latter
residual is $-28.7$ points, with a paired-problem 95\% interval of
$[-45.2,-13.4]$. Gemma-4-26B-A4B-it at 50\% expert pruning also falls
well below its reference, at 20.0\% versus 84.0\%. For both models with
large deficits, the shortfall is larger on LiveCodeBench-v6 than on
HumanEval+. Mean accuracy loss alone therefore does not explain how
coverage depends on pruning structure and task.

To compute the reference from the observed pools, let
$J_{qk}\sim\operatorname{Hypergeom}(n_{0q},c_{0q},k)$ be the number
of original-model successes in a uniformly chosen $k$-candidate subset.
The probability that none survives retention is $(1-a_m)^{J_{qk}}$,
giving
\begin{equation}
U_m(k)=\frac{1}{|\mathcal Q|}\sum_q
\left[1-\Exp\big[(1-a_m)^{J_{qk}}\big]\right],\qquad
D_m(k)=P_m(k)-U_m(k).
\label{eq:thinning}
\end{equation}
At the full original budget $K=n_{0q}$, the expectation reduces to
$(1-a_m)^{c_{0q}}$. The reference leaves problem weights and budgets
unchanged and requires $0\le a_m\le1$; undefined ratios and values
above one are excluded. It cannot create a success on an originally
unsolved problem, and agreement between two all-zero curves is
uninformative.

Spreading a fixed number of successes across more problems can increase
coverage without changing pass@1, while concentrating them can lower it.
The residual measures the departure from uniform loss but does not
identify how success probabilities changed. We quantify uncertainty with
2,000 paired-problem bootstrap resamples, re-estimating $a_m$ in each
and excluding resamples for which the reference is undefined. These
intervals describe variation across problems conditional on the observed
candidate pools.

Probability scaling gives ``uniform'' loss a particular meaning. To test
whether the coverage deficit persists under alternative definitions,
we use odds scaling,
$f_b(p)=bp/(1-p+bp)$, or hazard scaling, $f_b(p)=1-(1-p)^b$, with
$0<b\le1$ chosen to match mean accuracy. For empirical original
frequency $p>0$, each success is retained with probability $f_b(p)/p$
in the same finite-pool calculation; zero frequencies remain zero.
Both transformations preserve frequency one, so always-successful
problems impose a floor on the attainable mean. Targets below that
floor are infeasible; a target at the floor is a limiting case unless
the identity already attains it. No pseudocounts or clipping are used,
and retaining no successes is treated separately.

For Qwen3.6-35B-A3B FFN-40\% on AIME, coverage remains below these
alternative references, by 12.56 points under odds scaling and 25.69
under hazard scaling. The size of the deficit depends on the reference,
but its direction does not in this comparison.

\subsection{Changes in problem difficulty ordering}
\label{app:rank}
\label{sec:rank}
\label{app:rank-overview}
\label{app:ordering-support}
\label{app:certificate-validity}
\label{app:monotone-mismatch}
\label{app:all-budget-check}
\label{app:certificate-family}
\label{app:chunk-dependence}
\label{app:curve-information}

Even a close coverage fit can hide a different ordering of problem
difficulty. Swapping success probabilities $(0.8,0.2)$ to $(0.2,0.8)$,
for example, preserves IID coverage at every budget while reversing
which problem is easier. Permuting finite-pool success counts has the
same invariance when candidate counts are equal. We therefore ask whether
any common nondecreasing calibration $p_{mq}=f_m(p_{0q})$ can explain
the pruned probabilities, a stronger requirement than matching their mean.

\paragraph{Similar curves can conceal reordered difficulty.}
On AIME-2026, we use simultaneous probability intervals to identify pairs whose
relative difficulty reverses after pruning. A certified reversal
rules out a common nondecreasing calibration and gives a lower bound
on its error. We also check whether the same reversal remains certified
using all 2,048 responses, the first 256 and the remaining 1,792.

The Qwen3.6-35B-A3B and Gemma-4-26B-A4B-it comparisons show why
aggregate curve fit and evidence about difficulty ordering must be
assessed separately. Qwen3.6-35B-A3B FFN-5\% stays within
\numQwenFFNFiveMaxResidual{} points of the uniform-loss coverage curve
at every integer budget up to 2,048. Under IID sampling, however,
simultaneous intervals from the full pool imply that any monotone
calibration misses at least one problem by
\mismatchQwenFFNFiveIIDSup{} points. This full-pool result remains
exploratory because no reversed pair is certified in all three
windows. For Gemma-4-26B-A4B-it at 20\% expert pruning, the coverage
curve fits the uniform-loss reference less closely, with a maximum
residual of \numGemmaTwentyMaxResidual{} points. In this case,
difficulty reversals in \numRankGemmaTwentyStable{} problem pairs
remain certified across all three windows. Separately, under the IID
assumption, its full-pool intervals imply that any monotone calibration
errs by at least \mismatchGemmaTwentyIIDSup{} points on at least one
problem. Agreement in aggregate coverage therefore does not imply
preserved difficulty ordering.

\paragraph{Certifying reversals and calibration error.}
A certificate requires simultaneous intervals $[L_{tq},U_{tq}]$ for
the original ($t=0$) and pruned ($t=m$) success probabilities. A pair
of problems reverses order whenever
\begin{equation}
L_{0i}>U_{0j}\quad\text{and}\quad U_{mi}<L_{mj}.
\label{eq:rank-certificate}
\end{equation}
On the simultaneous coverage event, the original model finds $i$
easier than $j$, while the pruned model finds it harder. A common
nondecreasing calibration cannot reproduce both inequalities. Its
absolute errors on the pair must sum to at least the certified gap,
so at least one error is half that gap or larger. For the class
$\mathcal F$ of such functions and the certified pair set $\mathcal E$,
\begin{equation}
\Delta_\infty=\inf_{f\in\mathcal F}\max_q|p_{mq}-f(p_{0q})|
\ \ge\ \frac12\max_{(i,j)\in\mathcal E}(L_{mj}-U_{mi}).
\label{eq:monotone-mismatch}
\end{equation}
Because all intervals hold on the same event, selecting the largest
gap needs no further correction. We round bounds down, and an empty
pair set is inconclusive.

Under IID sampling within each prespecified stream, we use
Clopper--Pearson intervals with Bonferroni level $\eta=0.05/M$,
where $M$ counts all prespecified model--problem--window intervals,
with each shared original counted only once. Outcome-selected
results enter the correction but are excluded from certification claims.
The union bound controls the probability of any false certificate at
5\% without requiring independence across models, problems or windows.
Recurrence across overlapping windows checks stability without
providing independent replication or validating IID sampling. Aggregate
curve residuals are evaluated at every integer budget, so their reported
maxima do not depend on a sparse plotting grid.

\paragraph{How candidate dependence changes the evidence.}
To assess how the preceding results depend on the IID sampling
assumption, we treat each group of 32 candidates as one bounded
observation $X_b$, its success fraction. This permits arbitrary
dependence within a group while retaining the requirement of a
constant conditional mean across groups,
$\Exp[X_b\mid\mathcal F_{b-1}]=\mu$. Under this assumption, the full
2,048-response pool for Gemma-4-26B-A4B-it expert-20\% still yields
\numChunkGemmaTwenty{} certified pairs and a
\mismatchGemmaTwentyChunkSup{}-point calibration error bound. This
evidence does not persist in either shorter window, neither of which
certifies a pair. The certificates obtained under IID sampling also
disappear for Qwen3.6-35B-A3B FFN-5\% and GLM-4.7-Flash expert-50\%.
Which reversals can be certified thus depends on the assumptions
about dependence among candidates.

The group analysis inverts the following mixture
of betting products \citep{waudbysmith2024betting}, with
$\mathcal C=\{0.1,0.25,0.5,0.75,0.9,1\}$,
\begin{equation}
E_R^+(\mu)=\frac{1}{6}\sum_{c\in\mathcal C}
\prod_{b=1}^{R}\left(1-c+\frac{cX_b}{\mu}\right),\qquad
E_R^-(\mu)=E_R^+(1-\mu;1-X_1,\ldots,1-X_R).
\label{eq:chunk-betting}
\end{equation}
Each product is nonnegative with expectation one for $0<\mu<1$,
so Markov's inequality gives a tail bound. The lower endpoint is zero
for an all-zero stream, and the upper endpoint is one for an all-one
stream, with the opposite endpoint obtained by inversion. We compare
betting, Hoeffding and Bernoulli-KL bounds, allocating $0.05/3$ to
each over the same $M$ intervals. The betting tail threshold is
$6M/0.05$. Neither alternative certifies a full-pool reversal. The
IID and group analyses use separate error families, with no joint
95\% guarantee for selecting between them.

Failure to certify a reversal does not establish preserved ordering.
In particular, the eight-group prefix is too short at this correction
level. Its betting products at $\mu=1/2$ are bounded by $2^8$, below
the threshold, so every interval contains $1/2$. The bounds apply to
mean success under the stated dependence assumption, not to population
pass@$k$ under dependent sampling.

\subsection{Sensitivity to decoding and trajectory-length limits}
\label{sec:decoding}

We vary temperature, output caps and repetition penalties to examine
their effects on coverage and generation behavior. A total-output cap
limits both reasoning and the final answer without directly controlling
thinking duration \citep{muennighoff2025s1}.

\begin{figure}[!htbp]
\centering
\includegraphics[width=\textwidth]{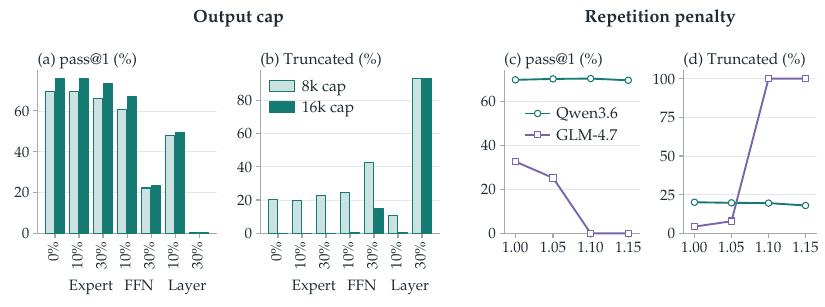}
\caption{\textbf{Output cap and repetition penalty.} Panels (a,b) compare
Qwen3.6-35B-A3B AIME accuracy and truncation under 8k/16k caps;
0\% denotes the unpruned model. Panels (c,d) vary repetition penalties
for the original Qwen3.6-35B-A3B and GLM-4.7-Flash models, labeled
Qwen3.6 and GLM-4.7, respectively.}
\label{fig:decodingcontrols}
\end{figure}

\paragraph{Larger output caps do not uniformly recover pruning losses.}
Whether more generation helps depends on what limits success. Raising
the output cap from 8{,}192 to 16{,}384 tokens benefits Qwen3.6-35B-A3B
under 30\% expert pruning, where AIME pass@1 rises from 65.92\% to 73.62\%,
truncation nearly disappears, and coverage reaches all 30 problems at
$k=2048$ (Figure~\ref{fig:decodingcontrols}, panels a,b).
At 30\% FFN pruning, coverage likewise increases from 25 to 27 problems
(Table~\ref{tab:cap-aime2026}).

At 30\% layer pruning, raising the cap produces occasional new successes
without a comparable recovery in accuracy. Coverage increases from three to
five problems, but pass@1 rounds to 0.37\% under both caps as mean length nearly
doubles from 7,806 to 15,443 tokens. Truncation remains at 93.0\%,
compared with 92.9\% under the original cap. For this model,
allowing longer responses mostly extends unsuccessful attempts.
The contrast with expert pruning shows why truncation alone does not
predict the benefit of a higher output cap.

On LiveCodeBench-v6, 30\% FFN pruning shows nearly unchanged accuracy
and coverage at the common $k=256$, despite longer responses
(Table~\ref{tab:cap-lcb}). Generation conditions and scoring procedures
also differ, so this comparison does not isolate the output-cap effect.

\paragraph{Repetition-penalty sensitivity.}
Repetition penalties affect the original models differently
(Figure~\ref{fig:decodingcontrols}, panels c,d). Across AIME-2026 repetition
penalties from 1.00 to 1.15, Qwen3.6-35B-A3B pass@1 ranges from 69.45\% to 70.22\%.
GLM-4.7-Flash pass@1 falls from 32.53\% to 25.14\% at 1.05 and to 0.00\% at
1.10, where truncation rounds to 100.0\%.

\paragraph{Temperature changes single-sample accuracy and coverage differently.}
We compare $T\in\{0.1,0.3,0.5,0.7,0.9\}$ at 50\% expert pruning,
using an 8,192-token cap and 2,048 candidates per AIME problem.
LiveCodeBench-v6 coverage is compared at $k=512$ for Qwen3.6-35B-A3B
and $k=256$ for GLM-4.7-Flash.

For GLM-4.7-Flash, raising temperature from
0.1 to 0.9 lowers AIME pass@1 from 21.19\% to 15.87\%, while
increasing coverage from 26 to 28 problems. On LiveCodeBench-v6,
pass@1 similarly falls from 24.69\% to 21.18\%, while coverage
rises from 48.57\% to 55.43\%. Qwen3.6-35B-A3B is less
sensitive in these sweeps: its AIME coverage ranges from 28 to 29
problems, and its LiveCodeBench-v6 coverage ranges from 81.71\%
to 84.57\% (Table~\ref{tab:temperature-controls}).

The GLM-4.7-Flash LiveCodeBench-v6 reference at $T=0.7$ uses a different scoring
procedure from the other temperatures, so differences involving this
reference cannot be attributed to temperature alone. The sweeps show
how decoder choice changes the accuracy--coverage tradeoff after pruning;
without corresponding original-model sweeps, they do not establish that
pruning itself creates a mismatch with the decoder.

\begin{table}[t]
\centering
\renewcommand{\arraystretch}{1.18}
{\small
\begin{tabular*}{\textwidth}{@{\extracolsep{\fill}}lrrr@{}}
\toprule
\textbf{Expert-50\%} & $T$ & \textbf{AIME-2026} & \textbf{LiveCodeBench-v6}\\
\midrule
Qwen3.6-35B-A3B & 0.1 & 64.71 / 96.67 & 49.03 / 81.71\\
Qwen3.6-35B-A3B & 0.3 & 64.77 / 96.67 & 49.04 / 82.86\\
Qwen3.6-35B-A3B & 0.5 & 63.96 / 93.33 & 49.18 / 82.29\\
Qwen3.6-35B-A3B & 0.7 & 63.53 / 96.67 & 49.05 / 84.57\\
Qwen3.6-35B-A3B & 0.9 & 62.54 / 96.67 & 48.78 / 84.57\\
\midrule
GLM-4.7-Flash & 0.1 & 21.19 / 86.67 & 24.69 / 48.57\\
GLM-4.7-Flash & 0.3 & 19.98 / 93.33 & 24.39 / 50.86\\
GLM-4.7-Flash & 0.5 & 18.67 / 90.00 & 23.91 / 50.86\\
GLM-4.7-Flash & 0.7 & 17.43 / 93.33 & 22.74 / 53.30\\
GLM-4.7-Flash & 0.9 & 15.87 / 93.33 & 21.18 / 55.43\\
\bottomrule
\end{tabular*}}
\caption{\textbf{Temperature sensitivity.} Cells give pass@1 / pass@$K$ (\%). AIME: $K=2048$. LiveCodeBench-v6: $K=512$ for Qwen3.6-35B-A3B and $256$ for GLM-4.7-Flash.}
\label{tab:temperature-controls}
\end{table}
\begin{table}[t]
\centering
\renewcommand{\arraystretch}{1.18}
{\small
\begin{tabular*}{\textwidth}{@{\extracolsep{\fill}}lrrrrrr@{}}
\toprule
\textbf{Configuration} & \textbf{8k scores} & \textbf{16k scores} & $\bar t_{8k}$ & $\bar t_{16k}$ & $\tau_{8k}$ & $\tau_{16k}$\\
\midrule
Original & 69.64 / 96.67 & 75.86 / 96.67 & 4923 & 5249 & 20.1 & 0.1\\
Expert-10\% & 69.51 / 96.67 & 75.75 / 96.67 & 4891 & 5231 & 20.0 & 0.1\\
Expert-30\% & 65.92 / 96.67 & 73.62 / 100.00 & 4845 & 5304 & 22.8 & 0.1\\
FFN-10\% & 60.56 / 96.67 & 66.83 / 100.00 & 5199 & 5640 & 24.4 & 0.2\\
FFN-30\% & 22.23 / 83.33 & 23.51 / 90.00 & 6118 & 7789 & 42.5 & 15.1\\
Layer-10\% & 48.08 / 93.33 & 49.45 / 93.33 & 5089 & 5235 & 10.5 & 0.4\\
Layer-30\% & 0.37 / 10.00 & 0.37 / 16.67 & 7806 & 15443 & 92.9 & 93.0\\
\bottomrule
\end{tabular*}}
\caption{\textbf{Output-cap controls on AIME-2026.} Scores are pass@1 / pass@2048 in per cent; $\bar t$ is mean output tokens and $\tau$ is truncation percentage.}
\label{tab:cap-aime2026}
\end{table}
\begin{table}[t]
\centering
\renewcommand{\arraystretch}{1.18}
{\small
\begin{tabular*}{\textwidth}{@{\extracolsep{\fill}}lrrrrrr@{}}
\toprule
\textbf{Configuration} & \textbf{8k scores} & \textbf{16k scores} & $\bar t_{8k}$ & $\bar t_{16k}$ & $\tau_{8k}$ & $\tau_{16k}$\\
\midrule
Original & 54.75 / 84.16 & 55.88 / 87.43 & 3489 & 3824 & 16.0 & 1.0\\
Expert-10\% & 54.99 / 85.04 & 56.27 / 88.00 & 3486 & 3836 & 16.1 & 1.1\\
Expert-30\% & 53.73 / 83.10 & 55.85 / 89.14 & 3484 & 3924 & 17.6 & 1.1\\
FFN-10\% & 48.11 / 81.49 & 50.28 / 84.57 & 4163 & 4882 & 26.2 & 2.7\\
FFN-30\% & 12.48 / 52.46 & 12.46 / 52.57 & 5261 & 8226 & 45.3 & 33.7\\
Layer-10\% & 35.03 / 69.51 & 36.04 / 72.00 & 3881 & 4595 & 21.3 & 3.9\\
Layer-30\% & 7.36 / 27.05 & 7.25 / 28.57 & 4360 & 8295 & 49.0 & 48.3\\
\bottomrule
\end{tabular*}}
\caption{\textbf{Output-cap comparisons on LiveCodeBench-v6.} Scores are pass@1 / pass@256 in per cent; $\bar t$ is mean output tokens and $\tau$ is truncation percentage.}
\label{tab:cap-lcb}
\end{table}

\section{Implications and Mixed-Axis Evaluation}
\label{app:discussion}
\label{app:budgets}
\label{sec:sampling-compensation}
\label{app:open-questions}

\subsection{Sampling allocation and cumulative coverage}
\label{app:allocation-transfer}
\label{app:allocation-blocks}
\label{app:allocation-controls}
\label{app:allocation-stopping}

The cost of rare successes motivates allocating more samples to problems
where they are most likely to add coverage. Yet a policy fitted to the
pruned model may perform well on a new batch simply by
re-solving problems already answered during calibration. We therefore
measure cumulative coverage across disjoint calibration and test
responses to the same AIME problems, then restrict allocation to those
unsolved during calibration.

\paragraph{Re-solving known problems can overstate the gain.}
The main split uses $t=256$ calibration responses and 1,792 held-out
responses per problem. Reversing the split tests sensitivity to
calibration depth. From the calibration success count $c^{\rm cal}_{jq}$, we estimate
$\widetilde p_{jq}=(c^{\rm cal}_{jq}+1/2)/(t+1)$ for problem $q$ and
original ($j=0$) or pruned ($j=m$) model. Half-count smoothing is used
only to plan allocation, while held-out coverage uses the unsmoothed
finite-pool estimator in Eq.~\ref{eq:passk}.

We compare a uniform policy (U), which assigns $b$ samples per problem,
with policies fitted to the original (O) or pruned (P) estimates, at
$b\in\{16,32,64,128\}$. Two controls examine whether the comparison
changes after adjusting the overall success level or restoring the
original problem ordering. Mean-adjusted original allocation (M) scales the original
estimates by $\widetilde a=\overline{\widetilde p}_m/
\overline{\widetilde p}_0$, matching average estimated success while
preserving relative probabilities. It is defined only for
$\widetilde a\le1$. Original-order allocation (R) assigns the sorted
pruned probability values to problems ranked by the original estimates.
Ties use increasing natural ID.

For any fitted probabilities $v_q$, allocation greedily takes the largest
remaining marginal gain $v_q(1-v_q)^{k_q}$. Diminishing gains make this
rule an exact maximizer of fitted coverage
$\sum_q[1-(1-v_q)^{k_q}]$ under the common integer budget
$\sum_q k_q=Qb$ and per-problem cap $n_{\rm test}$. Because R preserves
the probability multiset, its optimal fitted coverage equals P's.
Held-out coverage can nevertheless differ because successes remain
associated with their actual problems.

We evaluate each allocation in two ways. Test-batch coverage counts
only held-out successes, while cumulative coverage also counts problems
already solved during calibration. Writing $h_q(k)$ for the held-out
summand in Eq.~\ref{eq:passk}, with $h_q(0)=0$, and
$s_q=\mathbf{1}\{c^{\rm cal}_{mq}>0\}$, these are
$C_{\rm next}(A)=Q^{-1}\sum_q h_q(k_q^A)$ and
$C_{\rm incl}(A)=Q^{-1}\sum_q[s_q+(1-s_q)h_q(k_q^A)]$.
Both average over uniform subsets of the held-out pools. The pruned-model
budget is $Qt+Qb$, with a separate $Qt$ responses for original-model
calibration where used. All allocations are fixed from calibration
counts before held-out evaluation. U ignores labels, and O uses only
original-model labels.

This distinction changes the apparent benefit of recalibration. For
Qwen3.6-35B-A3B at 75\% expert pruning with $t=256,b=64$, P exceeds
M by 7.91 percentage points on the test batch but trails it by 1.00
point in cumulative coverage (Table~\ref{tab:allocation-examples}).
The advantage on the new batch therefore comes from re-solving
calibration successes, while fewer unsolved problems are reached.
The other examples likewise show why a test-batch gain alone does not
establish better use of the total sampling budget.

\begin{table}[!b]
\centering
\renewcommand{\arraystretch}{1.18}
{\small
\textbf{(a) Test-batch coverage and recalibration differences}\par\smallskip
\tabAllocationExamples
\par\medskip
\textbf{(b) P versus controls: coverage differences with and without calibration successes}\par\smallskip
\allocationControlExamples
\par\medskip
\textbf{(c) Cumulative coverage under unrestricted and restricted allocation}\par\smallskip
\allocationStoppingExamples
}
\caption{\textbf{Sampling-allocation comparisons on AIME-2026.} All panels
use $Q=30$, $t=256$, $b=64$ and 1,792 held-out candidates per problem.
Coverage is in per cent; differences are in percentage points. E denotes
expert pruning; $\Delta_M$ and $\Delta_R$ are P minus the corresponding
control. Next and Incl.\ denote test-batch and cumulative coverage,
Cal counts calibration-solved problems, and Used/limit reports samples
used and allowed. In (c), Unrestricted P allocates across all problems,
while P=U, O and M allocate only to problems unsolved during calibration.}
\label{tab:allocation-examples}
\end{table}

Differences between P and M reflect both changes in success probabilities
and estimation error. Differences between P and R also depend on how ties
are resolved, so they do not isolate ranking quality. We assess uncertainty with \numAllocationB{} paired-problem bootstrap
resamples, refitting probabilities, $\widetilde a$ and allocations.
The central 95\% ranges condition on the observed response pools and,
for M, on resamples satisfying $\widetilde a\le1$. Selected-budget
examples and the reversed split are exploratory comparisons on the
same problems, not independent replications.

Calibration itself also consumes samples. To separate the value of more
information from its cost, we vary $t\in\{32,64,128,256\}$ with the
same 1,792-response test pool, comparing a fixed subsequent budget
$b=64$ with a fixed total $t+b=320$. For GLM-4.7-Flash at 50\% expert
pruning, the cumulative P-versus-M ordering reverses with calibration
depth at fixed total budget. Recalibration thus has no uniform benefit
even before considering the cost of original-model responses.

\paragraph{Unsolved problems cannot be ranked by observed successes alone.}
Cumulative coverage gives no additional credit for re-solving a problem
already solved during calibration. We therefore restrict allocation to
$\mathcal U=\{q:c^{\rm cal}_{mq}=0\}$, the problems unsolved by the
pruned model. With allowance $B=Qb$ and common
cap $N=n_{\rm test}$, each policy uses
$B_{\rm used}=\min\{B,|\mathcal U|N\}$ samples on this set.
Coverage still uses all $Q$ problems, and pruned-model usage becomes
$Qt+B_{\rm used}$. Probability-based policies retain their full-set
fits, including M's scale, and allocate greedily within $\mathcal U$.
U divides the budget equally, assigning remainders by increasing ID.
Every policy, including O, now uses the pruned model's solved set.
The restriction is fixed before evaluation and uses no within-batch
feedback or stopping.

Every problem in $\mathcal U$ has the same pruned estimate,
$\widetilde p_{mq}=1/[2(t+1)]$, so calibration success counts provide
no basis for ranking these unsolved problems. With common caps and
ties, P consequently allocates exactly as U at every budget.
Restriction also changes the interpretation of R, since restricting
its vector need not preserve the probability multiset on the unsolved
subset. Skipping solved problems can still improve cumulative coverage
even when the remaining problems cannot be ranked. At fixed fits,
caps and ties, restriction weakly improves each policy's cumulative
coverage relative to its own unrestricted version, but this does not
establish an advantage over other policies. GLM-4.7-Flash at 50\%
expert pruning illustrates the distinction. Restricted P improves over
its unrestricted version, yet its allocation equals U and its
cumulative coverage remains below M.

\subsection{Model size and reasoning capacity under combined pruning}
\label{app:mixed-design}

The mixed-pruning comparison separates total model size from the
parameters used per token. Mixed-B is close in total size to 70\%
expert pruning of Qwen3.6-35B-A3B but has fewer active parameters
(Table~\ref{tab:mixed-parameters}). Comparing both quantities makes
the resource trade-off explicit, since matching total size does not
match active computation.

Mixed-A and Mixed-B combine 10\% layer and 50\% expert pruning with
20.31\% and 29.69\% FFN pruning, respectively
(Table~\ref{tab:mixed-parameters}). Both remove multi-token prediction
(MTP) modules and receive no recovery training. We construct them
through a common layer-pruning step followed by separate FFN and
expert selections, in that order. Each stage is non-iterative and
preserves earlier decisions. The common depth-pruned model is obtained
by selecting among ten disjoint, architecture-compatible four-layer
groups to minimize perplexity using 16 sequences of 2,048 tokens.
Starting from this model, each variant independently prunes routed
and shared FFNs using Taylor saliency with token-averaged signed
gradients on 32 sequences of 512 tokens. Expert selection then uses
RAZOR for each variant, retaining routing top-$k$ at eight.

\begin{table}[!htb]
\centering
{\small
\renewcommand{\arraystretch}{1.18}
\begin{tabular*}{\textwidth}{@{\extracolsep{\fill}}lrrrr@{}}
\toprule
\textbf{Model} & \textbf{AIME-2026} & \textbf{Minerva} & \textbf{HumanEval+} & \textbf{LiveCodeBench-v6}\\
\midrule
Expert-70\% & 51.23 / 96.67 & 47.94 / 87.50 & 85.59 / 99.39 & 46.17 / 77.98\\
Expert-75\% & 38.46 / 96.67 & 44.12 / 87.50 & 84.70 / 98.17 & 34.59 / 72.67\\
Expert-80\% & 13.14 / 90.00 & 31.74 / 86.40 & 77.68 / 97.56 & 23.50 / 58.85\\
\addlinespace[2pt]
Mixed-A & 23.65 / 83.33 & 35.67 / 84.93 & 65.23 / 96.95 & 21.15 / 52.00\\
Mixed-B & 16.68 / 83.33 & 28.19 / 83.09 & 53.40 / 92.68 & 12.03 / 40.00\\
\bottomrule
\end{tabular*}}
\caption{\textbf{Mixed-axis and expert-only pruning of Qwen3.6-35B-A3B.} Cells show pass@1 / pass@$K$ (\%): $K=2048$ for mathematics, $512$ for HumanEval+ and $256$ for LiveCodeBench-v6. Table~\ref{tab:mixed-parameters} defines the configurations; Appendix~\ref{app:sampling} details the Minerva grading differences.}
\label{tab:mixed-results}
\end{table}
\begin{table}[!htb]
\centering
{\small
\renewcommand{\arraystretch}{1.18}
\setlength{\tabcolsep}{3pt}
\begin{tabular*}{\textwidth}{@{\extracolsep{\fill}}lrrrrr@{}}
\toprule
\textbf{Configuration} & \textbf{Layers} & \textbf{Experts/layer} & \textbf{FFN width} & \textbf{Total} & \textbf{Active}\\
\midrule
Expert-70\% & 40 & 77 & 512 & 12.123 & 2.932\\
Expert-75\% & 40 & 64 & 512 & 10.486 & 2.931\\
Expert-80\% & 40 & 51 & 512 & 8.849 & 2.930\\
\addlinespace[2pt]
Mixed-A & 36 & 128 & 408 & 13.824 & 2.486\\
Mixed-B & 36 & 128 & 360 & 12.454 & 2.391\\
\bottomrule
\end{tabular*}}
\caption{\textbf{Parameter counts under mixed-axis pruning.} Total and active parameter counts are in billions and exclude vision and MTP modules, as in Table~\ref{tab:moe-coverage}. Active estimates the parameters used per token.}
\label{tab:mixed-parameters}
\end{table}

The solved sets clarify how the coverage deficit arises. On AIME at
$k=2048$, each mixed model's solved set is a strict subset of those
under 70\% and 75\% expert pruning. Relative to 80\% expert pruning,
each loses three problems and gains one. The deficit therefore reflects
both lost successes and a smaller number of newly observed successes,
rather than simply solving the same problems less frequently.

All total and active counts follow the definitions in
Appendix~\ref{app:operators}, excluding vision and MTP modules.
Original and mixed counts come from model weights, while expert-only
counts are estimated from the retained architecture. The structural
pruning ratios are nonadditive because the affected parameter groups
overlap, so the resulting total and active counts describe the
resources retained after the combined removals. Mixed-B's lower
observed coverage accompanies reduced active parameter use, but the
effect of this reduction on inference speed has not been measured.
The models also differ in MTP retention and pruning order, which
prevents attributing the coverage gap solely to how pruning is
distributed across structures.

\clearpage
\FloatBarrier
\makeatletter
\let\addcontentsline\appendix@addcontentsline
\makeatother
\end{document}